\documentclass[a4paper,fleqn]{cas-main}

\usepackage[authoryear,longnamesfirst]{natbib}
\usepackage{longtable}
\usepackage{caption}
\usepackage{graphicx}
\usepackage{tabularx}
\usepackage{booktabs}
\usepackage{subcaption}

\newif\ifdisplayabstract

\begin{document}
\let\WriteBookmarks\relax
\def\floatpagepagefraction{1}
\def\textpagefraction{.001}

\shorttitle{BrainIB++}

\shortauthors{T. Hu et~al.}

\title [mode = title]{BrainIB++: Leveraging Graph Neural Networks and Information Bottleneck for Functional Brain Biomarkers in Schizophrenia}                      

\author[1]{Tianzheng Hu}
\ead{tianzhenghu33@gmail.com}
\credit{Conceptualization, Methodology, Investigation, Formal analysis, Resources, Validation, Visualization, Writing – original draft}

\affiliation[1]{organization={Vrije University Amsterdam},
    addressline={De Boelelaan 1105}, 
    city={Amsterdam},
    postcode={1081 HV}, 
    country={The Netherlands}}

\author[2]{Qiang Li}
\ead{qli27@gsu.edu}
\credit{Methodology, Resources, Writing – review and editing}

\affiliation[2]{organization={Tri-institutional Center for Translational Research in Neuroimaging and Data Science (TReNDS), Georgia State, Georgia Tech, and Emory University},
    city={Atlanta, GA},
    postcode={30303}, 
    country={United States}}
    
\author[3]{Shu Liu}
\ead{liushu@mail.kiz.ac.cn}
\credit{Writing – review and editing}

\affiliation[3]{t={Key Laboratory of Genetic Evolution and Animal Models, Kunming Institute of Zoology, Chinese Academy of Sciences Kunming, China}
    }

\author[2]{Vince D. Calhoun}
\ead{vcalhoun@gsu.edu}
\credit{Writing – review and editing, Supervision}

\author[4]{Guido van Wingen}
\ead{g.a.vanwingen@amsterdamumc.nl}
\credit{Writing – review and editing, Supervision}

\affiliation[4]{organization={Department of Psychiatry, Amsterdam UMC, University of Amsterdam},
    city={Amsterdam},
    postcode={1100 DD}, 
    country={The Netherlands}}

\author[1,5]{Shujian Yu}
\cormark[1]
\ead{s.yu3@vu.nl}
\credit{Conceptualization, Methodology, Resources, Supervision, Writing – review and editing}

\affiliation[5]{organization={Department of Physics and Technology, UiT The Arctic University of Norway},
    city={Tromsø},
    country={Norway}}

\cortext[cor1]{Corresponding author}

\begin{abstract}
The development of diagnostic models is gaining traction in the field of psychiatric disorders.
Recently, machine learning classifiers based on resting-state functional magnetic resonance imaging (rs-fMRI) have been developed to identify brain biomarkers that differentiate psychiatric disorders from healthy controls. 
However, conventional machine learning-based diagnostic models often depend on extensive feature engineering, which introduces bias through manual intervention. While deep learning models are expected to operate without manual involvement, their lack of interpretability poses significant challenges in obtaining explainable and reliable brain biomarkers to support diagnostic decisions, ultimately limiting their clinical applicability. In this study, we introduce an end-to-end innovative graph neural network framework named BrainIB++, which applies the information bottleneck (IB) principle to identify the most informative data-driven brain regions as subgraphs during model training for interpretation. \textcolor{black}{We evaluate the performance of our model against nine established brain network classification methods across three multi-cohort schizophrenia datasets.}
It consistently demonstrates superior diagnostic accuracy and exhibits generalizability to unseen data. 
Furthermore, the subgraphs identified by our model also correspond with established clinical biomarkers in schizophrenia, particularly emphasizing abnormalities 
in the visual, sensorimotor, and higher cognition brain functional network. This alignment enhances the model's interpretability and underscores its relevance for real-world diagnostic applications. The code of our BrainIB++ is available at \url{https://github.com/TianzhengHU/BrainIB_coding/tree/main/BrainIB_GIB}.
\end{abstract}

 

\maketitle
\section{Introduction}
Schizophrenia is a mental disorder marked by recurrent episodes of psychosis, frequently involving distorted perceptions of reality. It affects approximately 24 million people or 1 in 300 people (0.32\%) worldwide \cite{saparia2022schizophrenia}. Common symptoms include hallucinations, delusions, disorganized thinking, social withdrawal, and flattened affect \cite{mueser1990hallucinations, bovet1993schizophrenic}.

The diagnosis of schizophrenia predominantly relies on clinical interviews with the Diagnostic and Statistical Manual of Mental Disorders\cite{edition2013diagnostic}.
Changes in brain structures, including gray matter, the cortex, and the striatum, have been extensively studied as possible biomarker candidates. These studies often integrate statistical analyses and mathematical modeling to elucidate the pathophysiology of psychiatric conditions.
Jiang et al. identified two distinct neurostructural subgroups by mapping the spatial and temporal trajectory of gray matter change in schizophrenia with the subtype and stage inference algorithm \cite{jiang2024neurostructural}.
MacKinley et al. leveraged cortical fold development in a time-locked fashion during fetal growth to investigate the timing of the prenatal insult linked to schizophrenia \cite{mackinley202s0deviant}.
Koshiyama et al. developed a novel multivector autoregressive modeling approach for assessing effective connectivity among cortical sources and found widespread patterns of hyper-connectivity across a distributed network of the frontal, temporal, and occipital brain regions \cite{koshiyama2020neurophysiologic}. 
Li et al. developed a new hypothesis-driven neuroimaging biomarker for schizophrenia identification, prognosis, and subtyping based on functional striatal abnormalities \cite{li2020neuroimaging}.

Functional changes are particularly relevant as they provide insights into brain activity and functional connectivity which have been shown to be disrupted by mental illness. By modeling the brain as a complex network of nodes and edges derived from resting-state functional magnetic resonance imaging (rs-fMRI), we can effectively study and assess alterations in whole-brain functional connectivity (FC) networks across diverse patient populations. In this context, differences in functional brain connectivity between patients with schizophrenia and healthy individuals reveal that the small-world properties of brain networks are more pronounced in schizophrenia, characterized by reduced FC strength in certain regions and increased FC diversity \cite{lee2019default, cai2022disrupted, lynall2010functional}.

In recent decades, classical machine learning methods have been extensively studied for schizophrenia diagnosis, enabling statistical inferences at the individual patient level \cite{arbabshirani2017single}. These techniques range from mathematically simple approaches, such as logistic regression, to more complex methods like ensemble learning techniques \cite{lei2020detecting, luo2020multimodal, xiao2019support}. However, a substantial portion of diagnostic research still relies heavily on feature engineering or manually crafted feature extraction \cite{shen2017using, tavakoli2024diagnosis}, utilizing metrics such as average strength, local efficiency, clustering coefficient, and transitivity \cite{mohammadkhanloo2024identification, algunaid2018schizophrenic}.
Even with expert knowledge, feature engineering can be ineffective or may inadvertently incorporate the biases of human experts during the design of handcrafted features. In contrast, deep learning, a specialized branch of machine learning, emphasizes training artificial neural networks with multiple hidden layers, allowing them to acquire hierarchical data representations while mitigating biases introduced by human intervention \cite{janiesch2021machine}.

The advancement of deep neural networks (DNNs) has been prominently demonstrated through the integration of fMRI data. For instance, Lin et al. employed spatial source phase maps derived from complex-valued fMRI data as inputs for a convolutional neural network (CNN)-based approach to classify schizophrenia \cite{lin2022sspnet}. Other examples include
CNN-LSTM hybrid model \cite{sarkar2024comparative, sujatha2021identification}, deep-learning based multiple sparsely connected network \cite{wang2024schizophrenia}, adversarial learning \cite{tang2023multi}.
Despite the effectiveness of DNNs, they may not fully leverage the non-Euclidean structure of brain networks.
Graph neural networks (GNNs), which combine deep learning with graph theory, have yielded promising results for analyzing these topological relationships \cite{xiang2020schizophrenia}. 
Graph convolutional networks (GCNs) effectively integrate node features with structural information for representation learning \cite{kipf2016semi}. For instance, Lei et al. applied GCNs to analyze topological abnormalities in functional brain networks in schizophrenia, achieving a classification accuracy of 85.8\% \cite{lei2022graph}. Chen et al. introduced a GCN method designed to learn brain network representations, effectively discriminating between patients with schizophrenia and normal controls \cite{chen2023classification}.
Expanding beyond GCNs, Lee et al. enhanced GCNs with a self-attention graph pooling method, improving graph classification performance \cite{lee2019self}. However, DNNs are often perceived as black boxes, rendering the interpretation of their results generally unreliable and potentially misleading the entire decision-making process of the model \cite{rudin2019stop}.
Consequently, enhancing the trustworthiness and interpretability of these approaches is essential to meet the stringent requirements of real-world medical applications \cite{kelly2019key, teng2022survey}.

To address this issue, numerous studies have been proposed, ranging from DNNs with DeepSHAP for estimating Shapley values \cite{reiter2020developing}, interpretable 3D-CNN framework incorporating saliency maps~\cite{lin2022sspnet}, Grad-CAM for enhanced model transparency \cite{selvaraju2017grad} to attention mechanisms \cite{choi2016retain, oktay2018attention, paschali2019deep, yang2024employing, miao2022interpretable} and ranking the top K strategy \cite{jiang2023assisting, chen2023discriminative, li2021braingnn, gao2023brain, kim2020understanding}.
Recently, mutual information has become popular in enhancing the model interpretability of GNNs. GNNExplainer maximizes the mutual information between predictions and subgraph structures to provide consistent explanations \cite{ying2019gnnexplainer}. 
The subgraph information bottleneck (SIB) framework, inspired by the information bottleneck (IB) principle \cite{tishby99information}, captures essential graph components while optimizing mutual information and addressing discrete data challenges through continuous subgraph selection \cite{yu2021recognizing}. However, the use of the mutual information neural estimator (MINE) \cite{belghazi2018mine} within SIB introduces high instability during training, which limits its effectiveness. Zheng et al. proposed BrainIB~\cite{zheng2022brainib} by replacing MINE with the matrix-based R\'enyi's $\alpha$-order mutual information \cite{yu2019multivariate} to stabilize the training. BrainIB has investigated the relationships among brain regions (referred to as edges in the brain graph) with stable training performance by employing IB on GCNs. Nevertheless, the subgraph generator in BrainIB presents significant challenges due to its complexity and lack of interpretability.

Our study builds upon BrainIB by introducing an enhanced version, BrainIB++. The key contributions of this work include:
\begin{itemize}
    \item \textbf{Methodological Improvements over BrainIB}
        \begin{itemize} 
            \item Rather than utilizing the physical automated anatomical labeling (AAL) parcellation map \cite{tzourio2002automated}, which does not naturally adapt to individual subject functional patterns, we applied a multi-model order template, which was derived using group-level independent component analysis (group ICA) from a large resting-state fMRI dataset of over 100K participants \cite{iraji2023identifying, duda2023reliability}.
            The group ICA-based parcellation is used as a spatial prior to estimate individual subject intrinsic connectivity networks and their functional network connectivity (FNC) rather than isolated brain regions, within the fully automated NeuroMark framework \cite{du2020neuromark}, addressing subject variability issues inherent in the AAL framework. 
            This approach enables the overcoming of subject variability problems in the AAL framework, facilitating more reliable comparisons across different subjects and enhancing the robustness of our findings in psychotic brain connectivity studies.
            \item In contrast to BrainIB, BrainIB++ samples subgraphs by selecting informative nodes, rather than edges. Specifically, BrainIB++ learns a node assignment matrix \( S \in \mathbb{R}^{n \times 2} \) (where \( n \) is the total number of nodes), with \( S_{i1} \) representing the probability that node \( i \) is within the subgraph, and \( S_{i2} = 1 - S_{i1} \) indicating the probability that node \( i \) is outside the subgraph. This sampling scheme significantly enhances the interpretability of subgraph generator.
        \end{itemize}
    \item \textbf{Generalization.} We evaluate BrainIB++’s performance against \textcolor{black}{nine} widely used brain network classification methods (including BrainIB) across two multi-cohort datasets, encompassing both single-cohort and multi-cohort experiments, with BrainIB++ consistently achieving superior diagnostic accuracy and generalizability. This strategy helps identify brain regions with the most abnormal functional connectivity to other areas, offering potential biomarkers for schizophrenia.
    \item \textbf{Built-in interpretability.} BrainIB++ effectively identifies subgraph biomarkers that not only align with clinical and neuroimaging findings but also demonstrate significant overlap with results obtained from traditional feature selection models.
\end{itemize}

\section{Background knowledge}

\subsection{Graph Neural Networks}
The graph is a type of structure that consists of brain regions, referred to as nodes, and their relationships, denoted as edges. GNN can process graphs to generate representations for downstream tasks, such as graph, node, or edge classification \cite{zhou2020graph}. Given a graph $\mathbf{G}=\{\mathcal{V};\mathcal{E}\}$. $\mathcal V = \{ V_i|i \in \{1,...,n \} \}$ denote nodes with node feature matrix \textcolor{black}{$X \in \mathbb{R}^{n \times d_X}$, $n$ is the number of nodes and $d_X$ denotes the dimension of features}. 
$\mathcal E = \{(V_i , V_j)|V_i, V_j \in \mathcal{V} \}$ denote the edges set of $\mathbf{G}$. $A = \{0, 1\} ^{n \times n}$ denote the adjacency matrix for the graph $\mathbf{G}$. 
The primary task of GNN is to efficiently propagate and aggregate information along the edges of the input graph using a message-passing mechanism, thereby acquiring the desired expressive representation.

Generalizing the convolution operator to irregular domains is typically achieved through neighborhood aggregation or a message-passing scheme.
This can be expressed mathematically as follows:
\begin{equation}
\mathbf{x}_i^{(k)} = \gamma^{(k)} \left(\mathbf{x}_i^{(k-1)}, \bigoplus_{j \in \mathcal{N}(i)} \phi^{(k)} \left( \mathbf{x}_i^{(k-1)}, \mathbf{x}_j^{(k-1)}, \mathbf{e}_{j,i} \right) \right),
\end{equation}
where $x_i^{(k-1)}$ denotes the features of node $V_i$ at previous layer $(k-1)$, $e_{j,i}$ denotes the edge features between node $V_j$ and $V_i$,  $\bigoplus$ denotes a differentiable, permutation invariant function, e.g., sum-, mean- or max-pooling \cite{kipf2016semi}, \cite{xu2018powerful}, \cite{hamilton2017inductive}.
$\phi^{(k)} (.)$ and $\gamma^{(k)} (.)$ are differentiable functions such as multi layer perceptron (MLP).
$\gamma^{(k)} (.)$ can also be defined as the aggregation function for the specific GNN.
For example, in a graph isomorphism network (GIN) using sum-pooling as the aggregation strategy, the message-passing procedure is expressed as follows:
\begin{equation}
    \mathbf{x}_i^{(k)} = \gamma^{(k)}\left( (1+\epsilon^k ) \cdot \mathbf{x}_{i}^{(k-1)} +  \sum_{j\in \mathcal{N}(i)} \mathbf{x}_j^{(k-1)}  \right),
\end{equation}
where $\epsilon$ denotes a learnable parameter and the initial $\mathbf{x}_0 = X$ in the first iteration.

In the context of graph classification, the representation of the whole graph $\mathbf{G}$ is obtained through a READOUT function, denoted as $\mathcal{R}$:

\begin{equation}
    X_{\mathbf{G}} = \mathcal{R} \left(\{ x_i^{(k)} | V_i \in \mathcal{V}  \} \right).
\end{equation}

Averaging and summation are among the most common strategies utilized for the READOUT function \cite{kipf2016semi}, \cite{xu2018powerful}. Another widely used strategy is hierarchical graph pooling, which reduces the number of nodes to one 
 \cite{ying2018hierarchical}, \cite{lee2019self}.

\subsection{Information Bottleneck Principle and its Extensions to Graph Data}

The IB principle is a technique for extracting information in one random variable $X$ that is relevant for predicting another random variable $Y$. IB operates by identifying a ``bottleneck" variable $\tilde{X}$ that maximizes its predictive power to $Y$, as expressed by the mutual information $I(\tilde{X};Y)$, while imposing some constraints on the amount of information it carries about $X$, formulated as the mutual information $I(\tilde{X};X)$:
\begin{equation}
    \mathcal{L}[p(\tilde{x}|x)] = I(\tilde{X}; X) - \beta I(\tilde{X}; Y),
\end{equation}
where $\beta $ is a Lagrange multiplier. By varying the only hyper-parameter $\beta$, one can explore the trade-off between the preserved meaningful information and compression at various resolutions \cite{tishby2000information}. 
By minimizing the $\mathcal{L}[p(\tilde{x}|x)]$, the IB principle also provides a natural approximation of minimal sufficient statistic \cite{gilad2003information}.




Several studies have introduced the IB principle to graph structure data, mainly in two areas, namely graph representation learning and graph interpretability. 
Specifically, the graph information bottleneck (GIB) framework enhances representation learning by optimizing the mutual information between the learned representation and the target, while minimizing it with respect to the input data \cite{wu2020graph}. \textcolor{black}{Specifically, GIB requires the node representation \textcolor{black}{$Z \in \mathbb{R}^{n \times d_Z}$} (where $n$ is the number of nodes and $d_Z$ is the dimension of the latent representation for each node)  to minimize the information from the graph-structured data $G$ and maximize the information to the node-level labels \textcolor{black}{$Y \in [K]^n$}, where each node is assigned one of $K$ classes}: 
\textcolor{black}{\begin{equation}
    \mathcal{L}_{\text{GIB}}  = \min I(G; Z) - \beta I(Y; Z).
\end{equation}}

On the other hand, Yu et al. ~\cite{yu2021recognizing} introduced the SIB framework, which focuses on extracting compressed, predictive subgraphs by reducing noise and redundancy while retaining key information. In contrast to GIB, the SIB aims to extract the most informative subgraph $G_{\text{sub}}$ from the input graph data $G$: 
\begin{equation}
\mathcal{L}_{\text{SIB}} = \min I(G; G_{\text{sub}}) - \beta I(Y; G_{\text{sub}}).
\end{equation}
The SIB can be regarded as a \textit{built-in} interpretable GNN, or a self-explaining GNN, as it automatically identifies the informative subgraph that has the greatest influence on the decision or graph label $Y$.
\textcolor{black}{The dynamic graph attention information bottleneck framework refines raw brain graphs by optimizing graph connections using the GIB to reduce noise and enhance feature aggregation with a Graph Attention Network (GAT) \cite{velivckovic2017graph} for improved brain network classification \cite{dong2024brain}. 
Additionally, the spatial-temporal dynamic hypergraph information bottleneck framework addresses challenges in automatic brain network classification by utilizing GIB for purifying graph structures and dynamically updating input signals, while integrating a hypergraph neural network and bi-directional long short-term memory to capture higher-order spatial-temporal associations, leading to better optimization and generalization in brain network analysis \cite{dong2024spatial}.}

\subsection{GNN Interpretability in Neuroscience}
Recently, there has been an increasing emphasis on the interpretability of GNNs in psychiatric diagnosis. Ranking the top K nodes and extracting subgraphs with information theory across brain regions have been commonly used methods for model interpretation in previous work.

In previous studies, ranking the top K nodes within brain regions has emerged as a popular method for explaining the model.
Jiang et al. selected multimodal clinical EEG features and demographic information as inputs for the graph neural network, subsequently ranking feature importance \cite{jiang2023assisting}.
Chen et al. also applied TopK pooling layers to a GCN model to highlight salient brain regions within multimodal features, enhancing the classification of schizophrenia at an individual level \cite{chen2023discriminative}.
BrainGNN utilizes a node pooling layer to group node clusters or retains the top K-ranked nodes as a subgraph; however, it has not been validated on additional atlases and continues to face challenges with feature extraction \cite{li2021braingnn}.
Gao et al. evaluated the interpretability of GNN models by directly ranking the average explanatory weight coefficients of brain regions \cite{gao2023brain}. Kim et al. developed a GIN model utilizing one-hot encoding to visualize the important regions of the brain \cite{kim2020understanding}. 

Mutual information has been proven to enhance the interpretability of graph neural networks recently. 
GNNExplainer is a widely used method for post-hoc explanations, generating consistent and concise interpretations for entire classes of instances while maximizing the mutual information between a GNN’s predictions and the distribution of possible subgraph structures \cite{ying2019gnnexplainer}.
Sunil\cite{sunil2024graph} and Gallo\cite{gallo2023functional} both utilized GNNExplainer for feature visualization and interpretability in their models. Miao et al. proposed Graph Stochastic Attention, which applies the IB principle to introduce stochasticity in attention weights, filtering out task-irrelevant graph components while learning reduced-stochasticity attention to select task-relevant subgraphs for interpretation \cite{miao2022interpretable}
The SIB framework, inspired by the IB principle, aims to eliminate noise and redundancy while extracting the interpretable components of the graph. It optimizes the challenges associated with mutual information and the discrete nature of graph data through continuous relaxation for subgraph selection, incorporating a connectivity loss for stabilization \cite{yu2021recognizing}. 

In a recent study, Zheng et al. introduced BrainIB, employing the IB principle on graph neural networks to identify the most informative edges for psychiatric diagnosis \cite{zheng2022brainib}. 
In this study, BrainIB++ provides multi-cohort analysis and built-in interpretability, yielding valuable insights into high-level biological and clinical applications for mental disorders.

\section{Methods}
\subsection{Estimating Subject-Specific ICNs and Their Time Courses in the Psychosis Dataset}
In this study, we used multi-objective optimization independent component analysis with reference (MOO-ICAR) to estimate subject-specific independent component networks (ICNs) and their time courses using the \textit{Neuromark\_fMRI\_2.1 network template}, which includes $105$ high-fidelity ICNs identified from over 100K subjects~\cite{iraji2023identifying}. These ICNs are organized into a large-scale network encompassing six major functional domains in humans: subcortical (SC), sensorimotor (SM), temporal (TEM), visual (VIS), higher cognition (HG), and cerebellar (CB). We employed the Group ICA of fMRI Toolbox (GIFT) v4.0c package (\url{http://trendscenter.org/software/gift}) to implement this technique \cite{iraji2021tools}. 

The MOO-ICAR algorithm, which employs spatial independent component analysis (scICA), optimizes two objective functions: one aimed at enhancing the overall independence of the networks and the other focused on improving the alignment of each subject-specific network with its corresponding template~\cite{du2020neuromark}. The two objective functions, \( J\left(S_l^k\right) \) and \( F\left(S_l^k\right) \), are presented in the following equation, which outlines how \textcolor{black}{to estimate the \( l^{th} \) network for }the \( k^{th} \) subject using the network template \( S_{l} \) as a guide:

\begin{equation}
\max \left\{\begin{array}{c}
J\left(S_l^k\right)=\left\{E\left[G\left(S_l^k\right)\right]-E[G(v)]\right\}^2 \\ \\
F\left(S_l^k\right)=E\left[S_l S_l^k\right]    
\end{array}     \text {s.t.}\left\|w_l^k\right\|=1 \right.
\end{equation}

Here, \( S_l^k = \left(w_l^k\right)^T \cdot X^k \) represents the estimated \( l^{th} \) network for the \( k^{th} \) subject, where \( X^k \) is the whitened fMRI data matrix for that subject, and \( w_l^k \) is the unmixing column vector to be determined through the optimization functions. The function \( J\left(S_l^k\right) \) optimizes the independence of \( S_l^k \) using negentropy. In this context, \( v \) is a Gaussian variable with mean zero and unit variance, \( G(.) \) is a nonquadratic function, and \( E[.] \) represents the expectation of the variable. Meanwhile, \( F\left(S_l^k\right) \) aims to enhance the correspondence between the template network \( S_l \) and the subject-specific network \( S_l^k \). The optimization problem is addressed by combining these two objective functions through a linear weighted sum, with each weight set to 0.5. Applying scICA with MOO-ICAR to each scan generates subject-specific ICNs for each of the \( I=105 \) network templates, along with their corresponding time courses.

\subsection{Graph Data Construction}
\begin{figure}[h]
    \centering
    \includegraphics[width=1\linewidth]{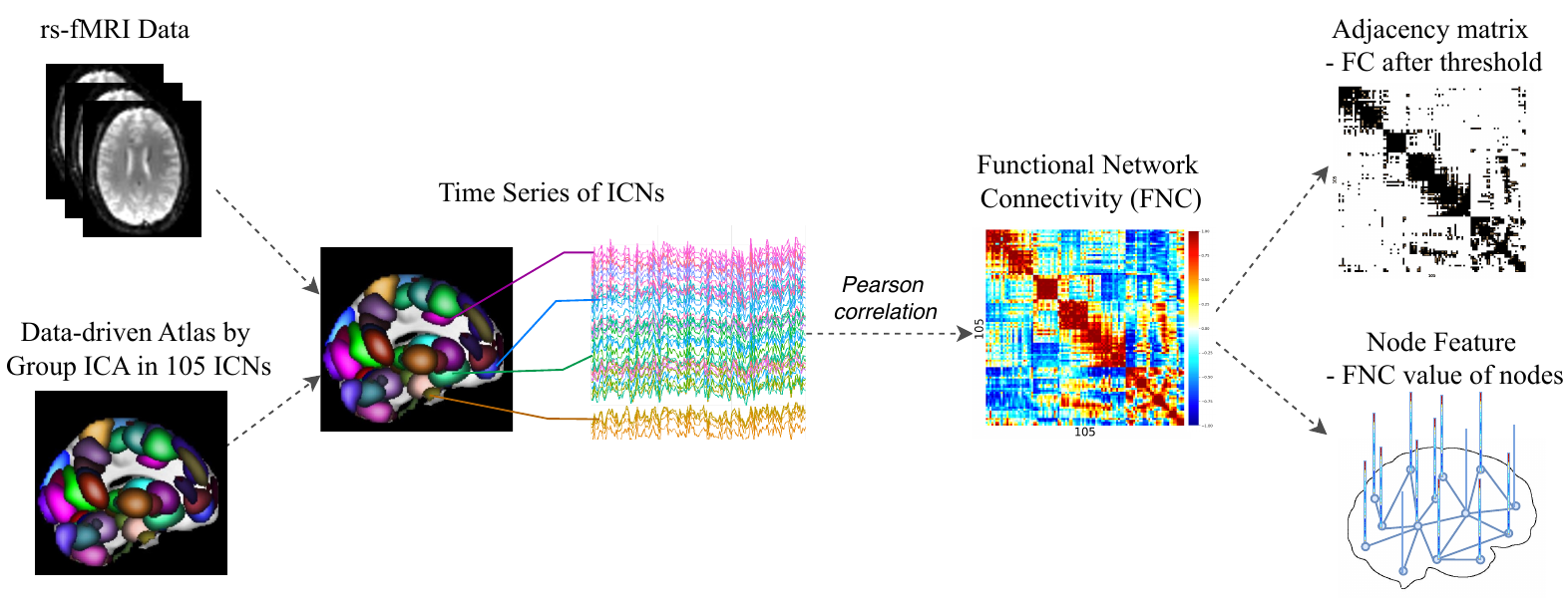}
    \caption{The pipeline to construct brain networks for graph neural networks from resting-state fMRI data. Data preprocessing of fMRI data involved preprocessing the raw resting-state fMRI data, followed by the application of spatially constrained group ICA using the \textit{NeuroMark\_fMRI\_2.1} template, which includes 105 ICNs to extract the corresponding time courses. The FNC matrices are computed using Pearson correlation between ICNs. From the FNC matrices, we construct the brain functional graph $G=\{X, A\}$. Specifically, $A$ is the adjacency matrix, where strong weighted functional connectivity values are binarized into ones between node pairs ($A \in \{0,1\}^{n \times n}$), while weaker connections are set to zero. $X$ is the node feature matrix with weighted functional connectivity values $(X \in \mathbb{R}^{n \times n})$. The weight feature of the $k$-th node $ X_k$ defined as $X_k = [\rho_{k1}, \rho_{k2},...,\rho_{kn}]^T$, where $n$ is the number of the nodes, and $\rho_{kl}$ represents Pearson correlation coefficient between node $k$ and node $l$.}
    \label{fig:data_preprocessing}
\end{figure}
    
Following preprocessing, time series data for 105 ICNs were obtained. Each region in the fMRI dataset corresponded to a specific brain functional network, enabling a thorough examination of functional connectivity. Functional network connectivity, computed using Fisher’s r-to-z transformed Pearson correlation between brain regions, resulted in a 105 $\times$ 105 symmetric functional connectivity matrix, which was subsequently used to generate a brain functional graph. Our analysis focused solely on positive connections to address concerns regarding the ambiguity associated with negative correlations. The FNC matrices were then used to derive feature vectors that encapsulate the connectivity profiles of individual brain regions. 
To transform the FNC matrices into adjacency matrices, a threshold of 0.4 was applied, with values above this threshold representing the presence of an edge between two brain regions. \textcolor{black}{While this threshold was initially established in our prior BrainIB work \cite{zheng2022brainib} and may not represent a universally optimal value, it was carefully chosen to retain 10–20\% of the strongest functional connections from 5,460 original edges, a range widely supported in neuroimaging literature for balancing noise reduction and signal preservation \cite{power2011functional, zalesky2012use, wang2023dynamic}.
To address potential bias, we evaluated thresholds from 0.2 to 0.5 on the BSNIP dataset (see Table \ref{tab:thresholds}). The results demonstrate that 0.4 achieves the highest performance (0.748), outperforming alternatives while maintaining a biologically plausible edge count (around 477 edges). Notably, lower thresholds (e.g., 0.2–0.3) retain excessive edges (more than 750), likely introducing noise. Higher thresholds (e.g., 0.5) oversparse the network (less than 300 edges), degrading discriminative power. 
In the context of this study, the 0.4 threshold ensures a balanced trade-off by retaining a sufficient number of edges to support effective graph-based modeling, while avoiding the inclusion of excessive spurious connections. However, although 0.4 is optimal for BSNIP, the best threshold may vary across cohorts due to site-specific noise. We recommend dataset-specific tuning due to phenotypic heterogeneity.}
For each node, the weights of the edges that connect it to other nodes served as its features. Ultimately, we generated an undirected weighted graph for each participant, comprising 105 nodes and edges with weights exceeding 0.4.

\begin{table}[width=.4\linewidth,cols=4,pos=h]
\centering
\caption{\textcolor{black}{Different threshold during functional connectivity construction on BSNIP dataset.}}
\label{tab:thresholds}
\setlength{\tabcolsep}{5pt}
\color{black}{
    \centering
    \begin{tabular}{ccc}
    \toprule
    \textbf{Threshold}  & \textbf{Performance} & \textbf{Avg. Edges}\\
    \midrule
     0.2 & 0.696 ± 0.041 & 1199.706\\
     0.25 & 0.702 ± 0.026 & 955.350\\
     0.3 & 0.705 ± 0.039 & 758.142\\
     0.35 & 0.717 ± 0.035 & 601.465\\
     0.4 & \textbf{0.748 ± 0.049} & 476.965 \\
     0.45 & 0.710 ± 0.036 & 378.206\\
     0.5 & 0.698 ± 0.060 & 299.201\\
    \bottomrule
    \end{tabular}
}
\end{table}

\subsection{Problem Definition}
Given a set of weighted brain networks $\{G^1, G^2,...,G^N\}$, the BrianIB++ model estimates the corresponding labels $\{y^1, y^2,..., y^N\}$, where $N$ is the number of participants. Define the $i$-th brain network $G^i=\{ X^i, A^i\}$. $A^i$ is the adjacency matrix. $X^i$ is the node feature matrix with weighted functional connectivity values $(X^i \in \mathbb{R}^{n \times n})$. Specifically, the weight feature of the $k$-th node $ X^i_k$ defined as $X^i_k = [\rho_{k1}, \rho_{k2},...,\rho_{kn}]^T$, where the $n$ is the number of the nodes, $\rho_{kl}$ is the Pearson’s correlation coefficient for node $k$ and node $l$. In Fig.\ref{fig:data_preprocessing}, we consider the functional connectivity values as node features and the binarized matrix as the adjacency matrix\cite{gallo2023functional}.

\subsection{Model Overview}
\begin{figure*}[h!]
    \centering
    \includegraphics[width=1\linewidth]{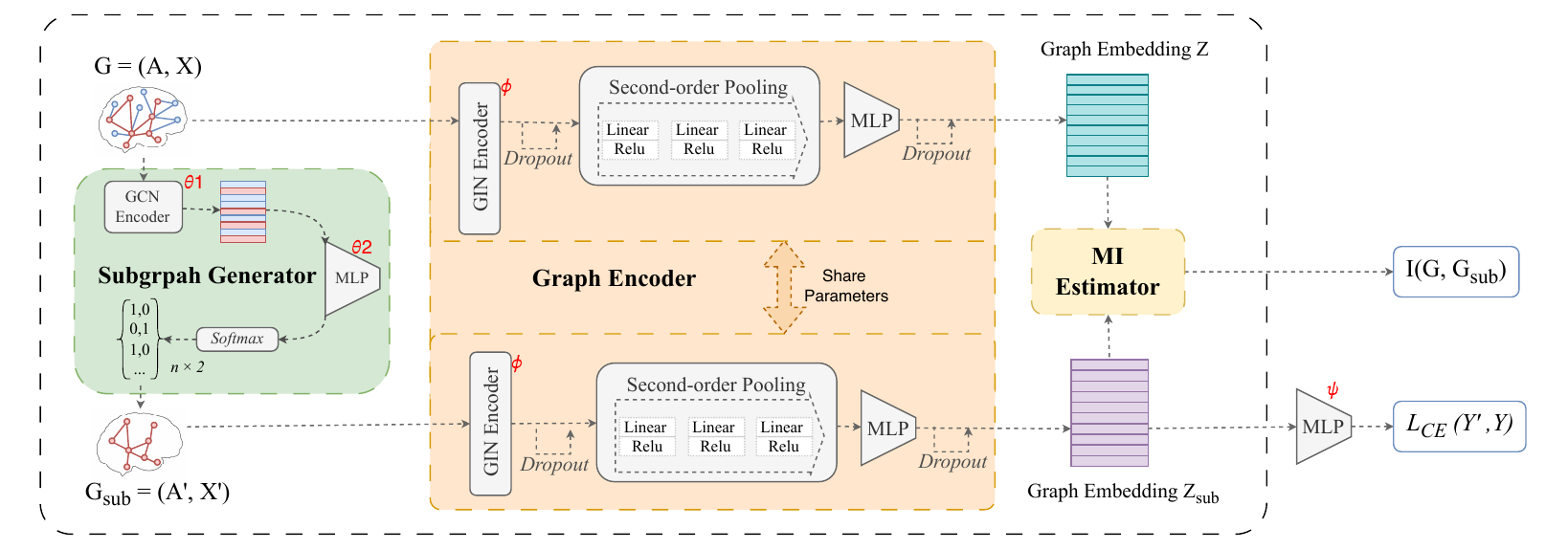}
    \caption{Architecture of our proposed BrainIB++. BrainIB++ consists of three modules including a subgraph generator, a graph encoder, and a mutual information estimator. Given the input graph data G, the subgraph generator samples subgraphs $G_{\text{sub}}$ along with the node assignment, indicating whether a specific node belongs to ${G}_{\text{sub}}$ or $\overline{{G}_{\text{sub}}}$. The graph encoder is used to learn graph embedding $Z$ and $Z_{\text{sub}}$ from $G$ or $G_{\text{sub}}$. The mutual information estimator assesses the mutual information between $I(G_{\text{sub}}, G)$ between $G$ and $G_{\text{sub}}$.}
    \label{fig:overview_framework}
\end{figure*}

BrainIB++ consists of three key modules: a subgraph generator, a graph encoder, and a mutual information estimator (see Fig. \ref{fig:overview_framework}).
The Subgraph Generator samples subgraphs $G_{\text{sub}}$ from the original graph $G$. The graph encoder extracts graph embeddings from either $G$ or $G_{\text{sub}}$. The mutual information estimator assesses the mutual information between $G$ and $G_{\text{sub}}$.

\subsubsection{Subgraph Generator}

Given the input graph structure $G$, a subgraph ${G}_{\text{sub}}$ with node assignments $S$ indicating whether a node belongs to ${G}_{\text{sub}}$ or its complement $\overline{{G}_{\text{sub}}}$ (see Fig. \ref{fig:overview_framework}). \textcolor{black}{
For instance, node $V_i$ corresponds to the $i$-the row of $S$, where $[\mathbb{P} (V_i \in G_{\text{sub}}),[\mathbb{P}(V_i \in \overline{G_{\text{sub}}})]$ represents the probabilities for node.
We initially use 2 layers GCN to obtain node embeddings, followed by 2 MLPs (2 layers fully connected network) to output assignment $S$:}
{\color{black} 
\begin{align}
 & H = \mathrm{GCN}_2(\text{relu}(\mathrm{GCN}_1(A,X)), \theta_1),\\
 & S = \text{softmax}(\mathrm{MLP}_2(\text{tanh}(\mathrm{MLP}_1(H)), \theta_2)).
\end{align}
}
\textcolor{black}{$H \in \mathbb{R}^{n \times d_H}$ is the node feature matrix after two GCN layers. }
$S$ is a matrix of dimensions $n\times 2$, where $n$ denotes the number of nodes. \textcolor{black}{This matrix represents the soft assignments of each node to latent subgraph components. The $\text{MLP}: \mathbb{R}^{d_H} \rightarrow \mathbb{R}^2$ reduce the} \textcolor{black}{dimension} \textcolor{black}{from ${d_H}$ to 2. Then a row-wise} softmax is applied to the output of the MLP, ensuring node assignment either in $\overline{G_{\text{sub}}}$ or ${G}_{\text{sub}}$. 
\textcolor{black}{
\begin{equation}
    S: \left\{
\begin{array}{l}
x_{1,1}, x_{1,2}, ... , x_{1,d_H}\\
x_{2,1}, x_{2,2}, ... , x_{2,d_H}\\
...\\
x_{n,1}, x_{n,2}, ... , x_{n,d_H}\\
\end{array}
\right\} \rightarrow 
\left\{
\begin{array}{l}
p_{1}, 1-p_{1}\\
p_{2}, 1-p_{2}\\
...\\
p_{n}, 1-p_{n}\\
\end{array}
\right\},
\end{equation}
where $p_i = \{\mathbb{P}(V_i \in G_{\text{sub}}) | i \in \{1,...,n\} \}$ is the possibility of the node $V_i$ belongs to $G_{sub}$.  When $S$ is optimally trained, the node assignments S are expected to converge to 0 or 1.} $\theta_1$ and $\theta_2$ are parameters of GCN and MLP, respectively.
\textcolor{black}{Then the assignment $S$ is used to calculate the CrossEntropy loss for the classification and the mutual information loss.}

\subsubsection{Graph Encoder}
The objective of the graph encoder is to encode the graph structure data ${G} = (X, A)$ into a vertical embedding (see Fig. \ref{fig:overview_framework}).
The encoder is applied to both the full graph data ${G}$ and the subgraph data ${G}_{\text{sub}}$ generated by the Subgraph Generator.

After two layers of the GIN, the representation $H$ of the original node feature matrix $X$ is obtained. Second-order pooling (SOPOOL) is applied to $H$, followed by dropout and normalization operations.
\textcolor{black}{\begin{equation}
    g_n = \mathrm{flatten} \left( \mathrm{SOPOOL}_{bi-linear} \left(GIN (A, X) \right), \phi \right),
\end{equation}
where $\phi$ is the parameter of the encoder.}

\subsubsection{Mutual Information Estimator}
The graph IB objective includes two mutual information terms:
\begin{equation}
    \underset{{G}_{\text{sub}}}{\mathrm{min}} I({G}_{\text{sub}}, {G}) - \beta I ({G}_{\text{sub}},Y).
\end{equation}
Minimizing $-I ({G}_{\text{sub}},Y)$ promotes ${G}_{\text{sub}}$ predictability of the graph label $Y$, which means maximizing the $I({G}_{\text{sub}}, Y)$:

\begin{align}
-I(G_{\text{sub}}, Y) &\leq \mathbb{E}_{Y,G_{\text{sub}}}  -\log q_\theta (Y | G_{\text{sub}}) \nonumber \\
&:= \mathcal{L}_{CE}(G_{\text{sub}}, Y)
\label{eq:mutual_info_bound},
\end{align}
where $q_\theta (Y | G_{\text{sub}})$ is the variational approximation to the true mapping $p (Y | G_{\text{sub}})$ from $G_{\text{sub}} $ to $Y$, and $\mathbb{E}_{Y,G_{\text{sub}}} $ denotes the expectation taken over the joint distribution of the label $Y$ and the subgraph $G_{\text{sub}}$. Eq. (\ref{eq:mutual_info_bound}) indicates that min $ - I (G_{\text{sub}}, Y )$ approximately equals to minimizing the cross-entropy loss $\mathcal{L}_{CE}$.

To compute the mutual information $I(G_{\text{sub}},G)$, we first vectorize the graph embeddings $Z$ and $Z_{\text{sub}}$ from $G $ and $G_{\text{sub}}$ respectively, using the Graph Encoder. We estimate $I(G_{\text{sub}},G)$ by evaluating $I(Z_{\text{sub}},Z)$, as the encoding process preserves sufficient information in $Z$ and $Z_{\text{sub}}$.
Use the matrix-based R\'enyi’s $\alpha$-order mutual information proposed by Yu et al~\cite{yu2019multivariate}, which is mathematically well-defined and computationally efficient to address this issue. 
This estimator is computationally efficient and does not require auxiliary neural networks, making it particularly suitable for deep learning applications.
Therefore, the matrix-based information $I({Z};{Z}_{\text{sub}})$ in analogy of Shannon’s could be expressed as:
\begin{equation}
    I({Z},{Z}_{\text{sub}}) = H_{\alpha}({Z}) + H_{\alpha}({Z}_{\text{sub}}) - H_{\alpha}({Z},{Z}_{\text{sub}}).
\end{equation}

Specifically, given a mini-batch of sample in size $M$, we have $\{Z^i\}_{i=1}^M$ and $\{Z^i_{\text{sub}}\}_{i=1}^M$ from graph encoder, where the $Z^i$ and $Z^i_{\text{sub}}$ indicate the $i$-th graph and the $i$-th subgraph in this mini-batch. We can evaluate the entropy of the graph embeddings from $\{Z^i\}_{i=1}^M$:

\begin{align}
H_{\alpha} (Z) &= \frac{1}{1 - \alpha} \log_2 \left( \operatorname{tr} \left( Q^{\alpha} \right) \right) \notag 
= \frac{1}{1 - \alpha} \log_2 \left( \sum_{i=1}^{N} \lambda_i (Q)^{\alpha} \right),
\end{align}
where the $tr$ denotes a trace of matrix, $Q = K / tr(K)$,xw $K = {k} \left(Z^i, Z^j \right)$ is the Gram matrix obtained from $\{Z^i\}_{i=1}^M$, $k$ is the positive definite kernel on all pairs of exemplars. $\lambda_i$ denotes the $i$-th eigenvalue of the $Q$.

Similarly, we can evaluate the entropy of subgraph embed-dings from $\{Z^i_{\text{sub}}\}_{i=1}^M$:
\begin{align}
H_{\alpha} (Z_{\text{sub}}) &= \frac{1}{1 - \alpha} \log_2 \left( \operatorname{tr} \left( Q_{\text{sub}}^{\alpha} \right) \right) \notag = \frac{1}{1 - \alpha} \log_2 \left( \sum_{i=1}^M \lambda_i (Q_{\text{sub}}^{\alpha}) \right),
\end{align}
where the $Q_{\text{sub}}$ also denotes the trace normalized Gram matrix evaluated on $\{Z^i_{\text{sub}}\}_{i=1}^M$ with the kernel function $k$.

Then the joint entropy between $Z$ and $Z_{\text{sub}}$ can be formalized as:
\begin{equation}
H_{\alpha} (Z, Z_{\text{sub}}) = H_{\alpha} \left( \frac{Q \circ Q_{\text{sub}}}{\operatorname{tr}(Q \circ Q_{\text{sub}})} \right),
\end{equation}
where the $Q \circ Q_{\text{sub}}$ denotes the Hadamard product between the $Q$ and $Q_{\text{sub}}$.

In which, the $Q$ and $Q_{\text{sub}}$ are obtained by the radial basis function (RBF) kernel $k$:

\begin{equation}
\kappa \left( z^i, z^j \right) = \exp \left( - \frac{\| z^i - z^j \|^2}{2\sigma^2} \right),
\end{equation}
where $\sigma$ denotes the width of kernel $k$, we estimate the 10 nearest distances of each sample obtain the mean value, and set the $\sigma$ with the average of mean values for all samples. The value of $\alpha$ is decided during the Hyper-parameter tuning and fixed to 1.01 to approximate Shannon mutual information finally (see Table \ref{tab:hyper-parameters}).

\textcolor{black}{
The overall loss function is:
\begin{align}
    \mathcal{L}_{\theta, \phi, \psi} = \mathcal{L}_{CE} + \lambda \cdot I({Z},{Z}_{\text{sub}}).
\end{align}
}
\textcolor{black}{
The $\mathcal{L}_{CE}$ is the cross-entropy loss on the subgraph-enhanced classifier, and $I({Z},{Z}_{\text{sub}})$ is the contrastive mutual information loss encouraging invariance between original and subgraph representations. The $\theta, \phi, \psi$ are the parameters of the Subgraph Generator, Graph Encoder, and Classifier, respectively. $\lambda$ is the coefficient that controls the weight or importance of the mutual information loss.}

\section{Experiment}
\subsection{Real-world Datasets and Preprocessing}
In this study, we utilized \textcolor{black}{three datasets: a dataset from the Bipolar and Schizophrenia Network for Intermediate Phenotypes (BSNIP) \cite{tamminga2013clinical, tamminga2014bipolar}, a dataset acquired from the UCLA Consortium for Neuropsychiatric Phenomics (UCLA) \cite{poldrack2016phenome}}, \textcolor{black}{and a dataset from the Center for Biomedical Research Excellence (COBRE) \cite{cobre_dataset}, see Table \ref{tab:summary_dataset}. 
The BSNIP dataset included sites in Baltimore, Chicago, Dallas, Detroit, Boston, Georgia, and Hartford. 
It }contains 1,142 individuals, including 640 control subjects (261 male and 379 female; 51 from Baltimore, 149 from Chicago, 118 from Dallas, 22 from Detroit, 35 from Boston, 104 from Georgia, and 161 from Hartford) and 502 subjects diagnosed with schizophrenia (322 male and 180 female; 69 from Baltimore, 90 from Chicago, 79 from Dallas, 23 from Detroit, 57 from Boston, 48 from Georgia, and 136 from Hartford), with ages ranging from 15 to 60 years old.
The UCLA dataset includes neuroimaging data acquired using two 3T Siemens Trio scanners. These scanners were located at the Ahmanson-Lovelace Brain Mapping Center (Siemens version syngo MR B15) and the Staglin Center for Cognitive Neuroscience (Siemens version syngo MR B17) at UCLA. The dataset consists of 50 subjects diagnosed with schizophrenia (8 female, 42 male) and 114 healthy control subjects (56 female, 58 male), with ages ranging from 21 to 50 years old.
\textcolor{black}{The COBRE dataset contains 72 patients with schizophrenia and 74 healthy controls, with ages ranging from 18 and 65 years. In this study, we utilized a publicly available subset of the COBRE dataset, which includes data from 47 schizophrenia patients (19 female, 28 male) and 34 controls (9 female and 25 male) \cite{liao2019preservation}. This subset is available through the freely accessible OpenNeuro \cite{openneuro2017} platform and has been used in previous studies for schizophrenia-related analysis \cite{lei2022graph}. Although it does not cover the entire dataset, this subset maintains representative samples from both groups and follows the same acquisition protocols, making it suitable for evaluating the generalizability of our model on independent data.}
\begin{table}[width=.55\linewidth,cols=4,pos=h]
\centering
\caption{\textcolor{black}{Summary of the dataset information. HC refers to healthy control participants, and SZ refers to participants diagnosed with schizophrenia.}}
\label{tab:summary_dataset}
{\color{black} 
\begin{tabular}{cclcc}
\toprule
\textbf{Dataset} & \textbf{Group} & \textbf{Age(std)} & \textbf{Gender(F/M)} &\textbf{Total}\\ 
\midrule
\multirow{2}{*}{BSNIP} 
    & SZ & 36.60 $\pm$ 12.08 & 180/322 & 502\\
    & HC &  35.83 $\pm$ 12.29  &  374/259   &  633 \\
\midrule
\multirow{2}{*}{UCLA} 
    & SZ &  39.17 $\pm$ 8.59 &38/12 & 50 \\
    & HC &   31.20 $\pm$ 8.76  &  56/54   &  110 \\
\midrule
\multirow{2}{*}{COBRE} 
    & SZ &  28.64 $\pm$ 9.08 & 19/28  & 47  \\
    & HC &   34.94 $\pm$ 9.71  & 9/25   &  34  \\
\bottomrule
\end{tabular}
}
\end{table}

\textcolor{black}{All three }datasets underwent preprocessing steps, which included rigid body motion correction, slice timing correction, and distortion correction, utilizing the FMRIB Software Library (FSL v6.0, \url {https://fsl.fmrib.ox.ac.uk/fsl/fslwiki/}) and the Statistical Parametric Mapping (SPM12, \url{http://www.fil.ion.ucl.ac.uk/spm/}) toolboxes within MATLAB. Initial volumes were discarded, and slice timing and head motion corrections were applied. Distortion correction, to address warps from susceptibility-induced distortions using multiple scans with varying acquisition parameters, was performed with the top-up toolbox in FSL with default settings. The fMRI images were then registered to the Montreal Neurological Institute template for standardization to a common space. Finally, the data were resampled to 3 \textit{mm}$^3$ isotropic voxels and spatially smoothed with a Gaussian kernel of 6 \textit{mm} full width at half maximum.

Subsequently, the spatially constrained group-level ICA, MOO-ICAR, was employed using the \textit{NeuroMark\_fMRI\_2.1 template} as a spatial prior to extract subject-level spatial networks and related time series.

\subsection{Baselines}
To demonstrate the capability and performance of BrainIB++, we compared it against seven widely recognized traditional machine learning and deep learning models on \textcolor{black}{three psychiatric datasets: BSNIP, UCLA, and COBRE}. The selected models include four traditional classifiers: support vector machine (SVM), K nearest neighbors (KNN), decision tree, and AdaBoost; and \textcolor{black}{five representative GNNs: GCN \cite{kipf2016semi}, GIN \cite{xu2019powerful}, GAT \cite{velickovic2018graph} and the Graph Transformer \cite{shi2020masked} as well as the baseline BrainIB \cite{zheng2022brainib}}.
Our BrainIB++ model prioritizes nodes rather than edges, in contrast to BrainIB and traditional machine learning models. This node-centric approach directly emphasizes the significance of individual brain regions rather than their interconnections.

All the models are trained on the graph structure data with functional correlation features except the SVM. 
We generated hand-crafted features instead for SVM model training and post-hoc interpretability shapley explanations analysis.
The hand-crafted graph measures are categorized into two broad types: nodal and global features. Global or local features give information about node connectivity and the importance of certain nodes in the brain. 8 nodal features are generated for each of the 105 ICNs per subject, which gives $8 \times 105 = 840$ features per subject. The eight nodal features included: edges, degree centrality, betweenness centrality, closeness centrality, load centrality, average shortest path length, degree, and average neighbor degree. Additionally, two global features (average shortest path length and average degree) were computed for each subject to capture overall graph characteristics.

\subsection{Experimental Setup}
\textcolor{black}{For each traditional method, we conducted a comprehensive grid search to optimize its performance. Table \ref{tab:traditional_hyperparams} presents the key hyperparameters for each traditional model, including SVM, Decision Tree, KNN, and AdaBoost.}

\begin{table}[width=.85\linewidth,cols=4,pos=h]
\centering
\caption{\textcolor{black}{Hyperparameter search space for traditional machine learning models}}
\label{tab:traditional_hyperparams}
{\color{black} 
\begin{tabular}{ccccc}
\toprule
\textbf{Model} & \textbf{Hyper-parameter} & \textbf{Range examined} & \textbf{Final specification} \\ 
\midrule
\multirow{3}{*}{SVM} 
    & kernel & \texttt{[`linear', `rbf']} & `linear' \\
    & C      & \texttt{[0.1, 1, 10]} & 1 \\
    & gamma  & \texttt{[0.1, 1, 10]} &  0.1\\
\midrule
\multirow{4}{*}{Decision Tree}
    & max\_depth & \texttt{[5, 10, 15, None]} & 15 \\
    & min\_samples\_split & \texttt{[2, 5, 10]} &5 \\
    & min\_samples\_leaf  & \texttt{[1, 2, 4]}& 1 \\
    & criterion           & \texttt{[`gini', `entropy']} & `entropy'\\
\midrule
\multirow{3}{*}{KNN}
    & n\_neighbors & \texttt{[3, 5, 7, 9, 11]} & 11\\
    & weights & \texttt{[`uniform', `distance']}  & `uniform'\\
    & p & \texttt{[1, 2]} & 1\\
\midrule
\multirow{3}{*}{AdaBoost}
    & n\_estimators       & \texttt{[50, 100, 200, 300]} & 50 \\
    & learning\_rate  & \texttt{[0.01, 0.1, 1]} & 1 \\
    & base\_estimator\_max\_depth  & \texttt{[1, 2, 3]} & 1  \\
    
\bottomrule
\end{tabular}
}
\end{table}

We implemented and tested the BrainIB++ model using PyTorch 1.9.0 and PyTorch Geometric 2.0.3. \textcolor{black}{The model was trained for 100 epochs, with a learning rate decay gamma set to 0.9, applied every 5 steps}. \textcolor{black}{The matrix-based R\'enyi’s $\alpha$-order denoted by $\alpha$, is an internal variable within the objective function. In order to avoid overfitting, we employed the dropout mechanism with a dropout ratio of 0.5 and early stopping strategies during training to further mitigate overfitting risks. Moreover, all other deep learning baselines were followed by the same setting with BrainIB++.} 
\begin{table}[width=.7\linewidth,cols=4,pos=h]
    \centering
    \caption{Hyperparameters of BrainIB++ model.}
    \begin{tabular*}{\tblwidth}{@{} LLL@{} }
    \toprule
     \textbf{Hyper-parameter} & \textbf{Range examined} & \textbf{Final specification }\\ \midrule
        batch size & [16,32,64] & 32\\ 
        model learning rate & [1e-4, 2e-4, 5e-4] & 5e-4 \\
        subgraph model learning rate & [1e-4, 2e-4, 5e-4] & 2e-4 \\
        $\alpha$ value & [1.01,2,5] & 1.01\\
        \bottomrule
    \end{tabular*}
    \label{tab:hyper-parameters}
\end{table}

\textcolor{black}{
The hyper-parameters of BrainIB++ evaluated, along with the final configurations used to achieve the reported results, are summarized in Table \ref{tab:hyper-parameters}. These hyper-parameters were determined either through a grid search process or by following the optimal settings reported in prior research. 
We also evaluated GIN, GCN, and GAT as encoder layers in the Graph Encoder module, and GIN was chosen based on its overall performance.}
We implemented \textcolor{black}{stratified} 10-fold cross-validation to evaluate model performance during the training process.
This approach ensures a reliable assessment by dividing the dataset into ten subsets, iteratively using one subset for testing and the remaining nine for training, thereby ensuring robust evaluation. \textcolor{black}{In addition, stratified 10-fold cross-validation also preserves the original class distribution within each fold. This ensures that both training and testing sets maintain a consistent ratio of patients to controls, thereby reducing the risk of biased learning or evaluation caused by skewed distributions.}

\subsection{Single-cohort Dataset Training}
\textcolor{black}{We trained both baseline models and our BrainIB++ model on the BSNIP, UCLA, and COBRE datasets.} In this section, models are trained and evaluated on the same dataset.


\begin{table}[width=.7\linewidth,cols=4,pos=h]
    \centering
    \caption{\textcolor{black}{Experiment performance of all baseline models and BrainIB++ model on the single-cohort datasets, which means training and testing on the same dataset. }
    }
    \color{black}{
    \begin{tabular}{ccccccc}
        \toprule
        \multirow{2}{*}{\textbf{Single-cohort}}
        & \multicolumn{6}{c}{\textbf{Traditional Model}}\\ 
        \cline{2-7} \\[-0.2cm]
        & & \textbf{SVM} &
         \textbf{KNN} & \textbf{Decision Tree}  &  \textbf{AdaBoost}\\   
        \midrule
        \textbf{BSNIP} &
        & 0.611 
        &  0.651 & 0.568 & 0.625 \\
        \textbf{UCLA} &
        & 0.667
        & 0.664 & 0.607 & 0.727\\
        \textbf{COBRE} &
        & 0.750 & 0.731 & 0.712 & 0.625 \\ 
        \midrule
        \multirow{2}{*}{\textbf{Single-cohort}}
        & \multicolumn{4}{c}{\textbf{Graph Model}} & \multicolumn{2}{c}{\textbf{Graph IB Model}} \\ 
        \cline{2-7} \\[-0.2cm] &   
         \textbf{GCN}&\textbf{GIN} & \textbf{GAT} & \textbf{Graph Transformer} &\textbf{BrainIB}&
         \textbf{BrainIB++}  \\       
         \midrule
         \textbf{BSNIP} & 0.743 & 0.733 & 0.725 & 0.711 &  0.722 & \textbf{0.748} \\
         \textbf{UCLA} &0.737 & 0.769 & 0.762 & 0.771 &  0.770 & \textbf{0.783}  \\
         \textbf{COBRE} &
         0.759 &  0.773 & 0.761 &  0.823 & 0.833 & \textbf{0.847}\\
        \bottomrule
    \end{tabular}}
    \label{tab: single-cohort performance}
\end{table}

Table \ref{tab: single-cohort performance} compares the classification accuracy of traditional machine learning models and graph-based models on the \textcolor{black}{BSNIP,  UCLA, and COBRE datasets. The traditional models considered include SVM, decision tree, KNN, and AdaBoost. Among these, SVM achieves the highest accuracy with scores of 0.750 on the COBRE dataset and AdaBoost achieves 0.727 on the UCLA dataset. In contrast, the decision tree model shows the poorest performance, with accuracies of 0.568 on the BSNIP and 0.607 on the UCLA dataset.}
\textcolor{black}{Graph-based models, including GCN, GIN, GAT, Graph Transformer, BrainIB, and BrainIB++, consistently demonstrate superior performance compared to traditional models. Notably, the BrainIB++ model achieves the highest accuracy, with 0.748 on the BSNIP dataset, 0.783 on the UCLA dataset, and 0.847 on the COBRE dataset. The Graph transformer model also performs well, particularly on the smaller UCLA, and COBRE datasets, where it matches the performance of BrainIB and BrainIB++ with an accuracy of 0.771 and 0.823.}

\textcolor{black}{Training the SVM model with hand-crafted features results in lower performance compared to using the original graph data, achieving accuracies of 0.611 on the BSNIP dataset, 0.667 on the UCLA dataset, and 0.750 on the COBRE dataset.
In contrast, BrainIB++ consistently outperforms both traditional and graph-based models across all single-cohort datasets. Notably, it achieves an accuracy of 0.748 on BSNIP, surpassing the best-performing traditional model (KNN) by approximately 19.7\%. While the performance margin is narrower on the UCLA dataset, BrainIB++ still achieves the highest accuracy (0.783), demonstrating strong robustness. On the COBRE dataset, BrainIB++ reaches an accuracy of 0.847, closely matching the performance of advanced graph models like Graph Transformer (0.823), while maintaining model interpretability. These results highlight the superior performance and potential generalizability of BrainIB++ across diverse cohorts.}

\subsection{Multi-cohort Dataset Training}
In this experiment, aimed at evaluating the model’s generalizability, we trained and tested the models across different datasets. \textcolor{black}{This involved scenarios where the model was trained on the larger BSNIP dataset and tested on the smaller UCLA dataset or COBRE dataset.} This approach allows for a thorough assessment of the model's ability to generalize across different data sources.

\begin{table}[width=.8\linewidth,cols=4,pos=h]
    \centering
    \caption{\textcolor{black}{Experiment performance of all baseline models and BrainIB++ model on the multi-cohort datasets, which  means training and testing on different datasets. }
    }
    \color{black}{
    \begin{tabular}{ccccccc}
        \toprule
        \multirow{2}{*}{\textbf{Multi-cohort}}
        & \multicolumn{6}{c}{\textbf{Traditional Model}}\\ 
        \cline{2-7} \\[-0.2cm]
        & & \textbf{SVM} &
         \textbf{KNN} & \textbf{Decision Tree}  &  \textbf{AdaBoost}\\   
        \midrule
        \textbf{BSNIP $\rightarrow$ UCLA} &
        & 0.697 & 0.667 & 0.424 & 0.455 & \\
        \textbf{BSNIP $\rightarrow$ COBRE}&
        & 0.250 & 0.673 & 0.610 & 0.438 \\ 
        \midrule
        \multirow{2}{*}{\textbf{Multi-cohort}}
        & \multicolumn{4}{c}{\textbf{Graph Model}} & \multicolumn{2}{c}{\textbf{Graph IB Model}} \\ 
        \cline{2-7} \\[-0.2cm] &   
         \textbf{GCN}&\textbf{GIN} & \textbf{GAT} & \textbf{Graph Transformer} &\textbf{BrainIB}&
         \textbf{BrainIB++}  \\       
         \midrule
         \textbf{BSNIP $\rightarrow$ UCLA} & 0.565 & 0.571 & 0.594 & 0.568 &  0.693 & \textbf{0.709} \\
         \textbf{BSNIP $\rightarrow$ COBRE} & 0.584 & 0.632 & 0.621 & 0.646 & 0.667  & \textbf{0.681}\\
        \bottomrule
    \end{tabular}}
    \label{tab: multi-cohort performance}
\end{table}

Table \ref{tab: multi-cohort performance} presents the performance comparison of the baseline models and our BrainIB++ model.
In this scenario, the accuracies of most models declined during transfer learning on a multi-cohort dataset. The decline in accuracies can be attributed to the differences between in-distribution and out-of-distribution data. \textcolor{black}{ Specifically, the accuracies of GCN, GIN, GAT, and Graph Transformer decreased significantly by at least 15.3\%, 10.1\%, 10.4\%, and 6.5\% respectively. In contrast, BrainIB++ demonstrated much greater robustness, with a maximum accuracy drop of only 3.9\%, while still achieving notable accuracies of 0.709 and 0.681 on the UCLA and COBRE test datasets, respectively.}

\textcolor{black}{
When the models were trained on the BSNIP dataset and tested on the UCLA dataset, the traditional models, Decision Tree and AdaBoost performed poorly, achieving only 0.424 and 0.455 accuracy, respectively.} However, the accuracy of SVM and KNN slightly improved to 0.697 and 0.667, suggesting their potential adaptability in this setting.
This indicates that the hand-crafted features used in SVM are effective in capturing relevant patterns that generalize well across different datasets. 
\textcolor{black}{
In contrast, graph-based models demonstrated generally more stable multi-cohort performance. GCN, GIN, GAT, and Graph Transformer achieved accuracies ranging from 0.565 to 0.594, showing moderate improvements over traditional models, but still short of optimal generalization. Notably, BrainIB and BrainIB++ surpassed all baseline methods, with BrainIB achieving 0.693 accuracy and BrainIB++ reaching the highest performance of 0.709. These results emphasize the effectiveness of graph information bottleneck approaches in learning transferable representations across cohorts.}

\textcolor{black}{
When testing was performed on the COBRE dataset, a significant performance drop was observed for the traditional SVM model (0.250), while other traditional models showed more consistency (e.g., KNN at 0.673 and Decision Tree at 0.610). This suggests that SVM may be more sensitive to data distribution shifts between cohorts.}
\textcolor{black}{
In the graph-based category, GIN (0.632), GAT (0.621), and Graph Transformer (0.646) outperformed GCN (0.584) but still lagged behind the graph information bottleneck models. BrainIB achieved an accuracy of 0.667, which was the second highest among all models and scenarios, demonstrating robustness and generalization to unseen datasets. BrainIB++ achieved the highest accuracy of 0.681, surpassing all traditional and non-information bottleneck graph models, thereby highlighting the effectiveness of incorporating information bottleneck principles into graph learning.}

\subsection{Template Comparison of GroupICA and AAL}
\textcolor{black}{To assess the generalizability of our model across different brain parcellation schemes, we conducted additional experiments using the AAL template with 116 regions of interest (ROIs) on the BrainIB++ framework. Specifically, we tested GNNs including GCN, GIN and GAT on the UCLA dataset and the COBRE dataset, where the functional connectivity matrices were reconstructed based on the AAL template.}

\begin{table}[width=.7\linewidth,cols=4,pos=h]
\centering
\caption{\textcolor{black}{Comparison of Template AAL and groupICA.}}
\label{Comparison_AAL_groupICA}
{\color{black} 
\begin{tabular}{ccc|cc}
\toprule
\multirow{2}{*}{\textbf{Model}}&
\multicolumn{2}{c}{\textbf{UCLA}}&
\multicolumn{2}{c}{\textbf{COBRE}}\\
\cline{2-5}\\
[-0.2cm]
& \textbf{AAL\_116} & \textbf{groupICA\_105} & \textbf{AAL\_116} & \textbf{groupICA\_105} \\
\midrule
GCN & 0.706 ± 0.029 & \textbf{0.737 ± 0.049} & 0.758 ± 0.163 & \textbf{0.759 ± 0.143} \\
GIN & 0.730 ± 0.032 & \textbf{0.769 ± 0.051} & 0.709 ± 0.137 & \textbf{0.773 ± 0.145}\\
GAT & 0.700 ± 0.018 & \textbf{0.761 ± 0.051} & 0.746 ± 0.126 & \textbf{0.761 ± 0.116}\\
BrainIB++ &  0.694 ±  0.009  & \textbf{0.783 ± 0.077} & 0.597 ± 0.083 & \textbf{0.847 ± 0.084} \\
\bottomrule
\end{tabular}
}
\end{table}

\textcolor{black}{
Table \ref{Comparison_AAL_groupICA} shows the performance of each model under both the AAL\_116 and groupICA\_105 parcellation schemes. Across all models and datasets, results consistently demonstrate that the groupICA-based parcellation yields better classification performance than the traditional AAL atlas.}

\textcolor{black}{
On the UCLA dataset, all models achieved higher accuracy with groupICA\_105 compared to AAL\_116. For instance, GCN improved from 0.706 to 0.737, and GIN increased from 0.730 to 0.769. 
This trend also held clearly on the COBRE dataset. The performance differences between parcellation methods were pronounced, particularly for BrainIB++, which scored 0.597 with AAL\_116 but significantly improved to 0.847 with groupICA\_105 }

\textcolor{black}{These findings highlight that groupICA, as a data-driven parcellation method, captures more informative and discriminative brain connectivity patterns than the anatomically defined AAL template. The consistent advantage of groupICA across models and datasets suggests it is better suited for tasks involving functional connectivity-based brain disorder classification. Notably, the substantial performance improvement observed for BrainIB++ under the groupICA scheme indicates that, within our framework, the ICA-based method aligns particularly well with the principles of the information bottleneck theory, thereby facilitating more effective learning of task-relevant representations.}

\subsection{Ablation Study}
\textcolor{black}{We conducted additional ablation experiments to clarify the independent contributions of the node selection module. Specifically, we implemented an ablation baseline using four standard GNN models, including GCN, GIN, GAT, and Graph Transformer. The full BrainIB++ model includes the IB-based node selection module (see Table \ref{ablation_studies_single_cohort} and Table \ref{ablation_studies_multi_cohort}).}

\begin{table}[width=.8\linewidth,cols=4,pos=h]
\centering
\caption{\textcolor{black}{Model performances are compared under single-cohort scenarios, with and without the IB-based node selection module.}}
\label{ablation_studies_single_cohort}
{\color{black} 
\begin{tabular}{lccccc}
\toprule
        \multirow{2}{*}{\textbf{Module}}& \multirow{2}{*}{\textbf{Backbone}}&
        \multicolumn{3}{c}{\textbf{Single Cohort}}\\ 
        \cline{3-5} \\
        [-0.2cm]
        & & \textbf{BSNIP} & \textbf{UCLA} & \textbf{COBRE}\\ 
\midrule
\multirow{3}{*}{W/O. IB}
& GCN & 0.743 ± 0.042 & 0.737 ± 0.049 & 0.759 ± 0.143  \\
& GIN &0.733 ± 0.038 & 0.769 ± 0.051 &  0.773 ± 0.145     \\
& GAT &0.725 ± 0.013 & 0.762 ± 0.051 & 0.761 ± 0.116   \\
& Graph Transformer & 0.711 ± 0.015 &0.771 ± 0.018 & 0.823 ± 0.139 \\
\midrule
 \multirow{2}{*}
 {W. IB} 
 & BrainIB  & 0.722 ± 0.031  &   0.770 ± 0.053  &   0.833 ± 0.090    \\
& BrainIB++ & \textbf{0.748 ± 0.049} &  \textbf{0.783 ± 0.077} & \textbf{0.847 ± 0.084  }    \\
 
\bottomrule
\end{tabular}
}
\end{table}

\begin{table}[width=.7\linewidth,cols=4,pos=h]
\centering
\caption{\textcolor{black}{Model performances are compared under multi-cohort scenarios, with and without the IB-based node selection module.}}
\label{ablation_studies_multi_cohort}
{\color{black} 
\begin{tabular}{lcccc}
\toprule
        \multirow{2}{*}{\textbf{Module}}& \multirow{2}{*}{\textbf{Backbone}}&
        \multicolumn{2}{c}{\textbf{Multi Cohort}}\\ 
        \cline{3-4} \\
        [-0.2cm]
        &  & \textbf{ BSNIP $\rightarrow$ UCLA} & 
        \textbf{ BSNIP $\rightarrow$ COBRE}\\ 
\midrule
\multirow{3}{*}{W/O. IB}
& GCN & 0.565 ± 0.029 &  0.584 ± 0.145  \\
& GIN & 0.571 ± 0.070 & 0.632 ± 0.132    \\
& GAT & 0.594 ± 0.061  & 0.621 ± 0.165 \\
& Graph Transformer & 0.568 ± 0.136 & 0.646 ± 0.191 \\
\midrule
 \multirow{2}{*}
 {W. IB} 
 & BrainIB  & 0.693 ± 0.046 &  0.667 ± 0.063 \\
  & BrainIB++ & \textbf{0.709 ± 0.024}  &   \textbf{0.681 ± 0.178}      \\
\bottomrule
\end{tabular}
}
\end{table}

\textcolor{black}{The results in Table \ref{ablation_studies_single_cohort} (single-cohort setting) demonstrate that the addition of the IB node selection module consistently improves performance across all datasets. Specifically, the BrainIB++ model achieves the highest classification scores in all three cohorts (BSNIP: 0.748, UCLA: 0.783, COBRE: 0.847), outperforming all baselines by a notable margin. This indicates the superiority of the IB-enhanced representation in capturing discriminative patterns across individual datasets.}
\textcolor{black}{
The effectiveness of the IB node selection module becomes even more pronounced in the multi-cohort setting (Table \ref{ablation_studies_multi_cohort}).  BrainIB++ exhibits significantly stronger cross-cohort generalizability. 
These results clearly demonstrate that the proposed IB module not only enhances within-cohort learning but also substantially improves robustness and transferability in multi-cohort scenarios.}

\section{Interpretable Analysis}
\textcolor{black}{We primarily focus on the BSNIP and UCLA datasets for interpretability analysis, as BrainIB++ demonstrates superior performance under multi-cohort training on these datasets.
To facilitate visualization of the learned subgraph patterns, we align the spatial parcellation map derived from group ICA with the anatomically defined AAL parcellation map using BrainNet Viewer \cite{xia2013brainnet}.} BrainNet Viewer is an open-source, Matlab-based software package with a graphical user interface, that illustrates connectomes as ball-and-stick models. This alignment is achieved by calculating the shortest Euclidean distance in spatial space between each region in the group ICA map and each region in the AAL map, ensuring accurate correspondence between the two parcellation schemes. The Group ICA map contains 105 ICNs, while the AAL map contains 116; therefore, multiple regions from the Group ICA map may correspond to the same region in the AAL map. The detailed matching can be found in the \textit{supplementary file}. \textcolor{black}{Besides, in order to further evaluate the interpretability of our model, we conducted a quantitative comparison with a widely used post-hoc explanation method GNNExplainer \cite{ying2019gnnexplainer}. BrainIB++ surpasses GNNExplainer by achieving more reliable and intrinsically interpretable results, with full metric comparisons provided in the \textit{supplementary file}.}

\subsection{Traditional Methods}
Post-hoc interpretability analysis is commonly employed to explain the classification results after model training. 
Hand-crafted features assist the Shapley explainer by highlighting influential brain regions within the SVM model, ranking them according to their importance, and indicating their percentage contributions. This approach allows us to extract ranked importance biomarkers from the SVM model.

\begin{figure*}[h]
    \centering
    \begin{subfigure}[b]{0.45\linewidth}
    \centering
    \includegraphics[width=1\textwidth]{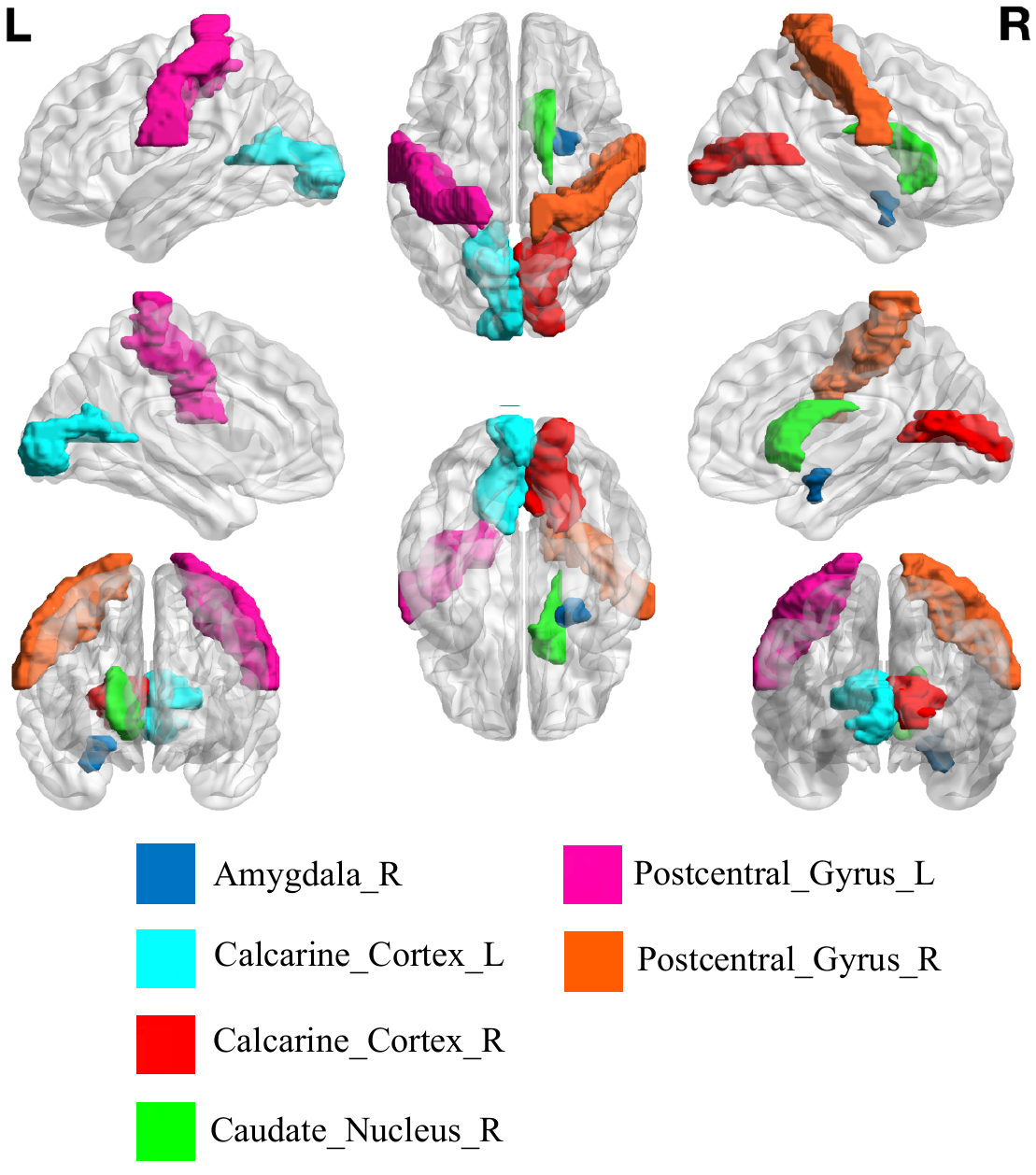}
    \caption{\textcolor{black}{single-cohort SVM}}
    \label{fig:single-cohort SVM}
    \end{subfigure}
    \hfill
    \begin{subfigure}[b]{0.45\linewidth}
    \centering
    \includegraphics[width=1\textwidth]{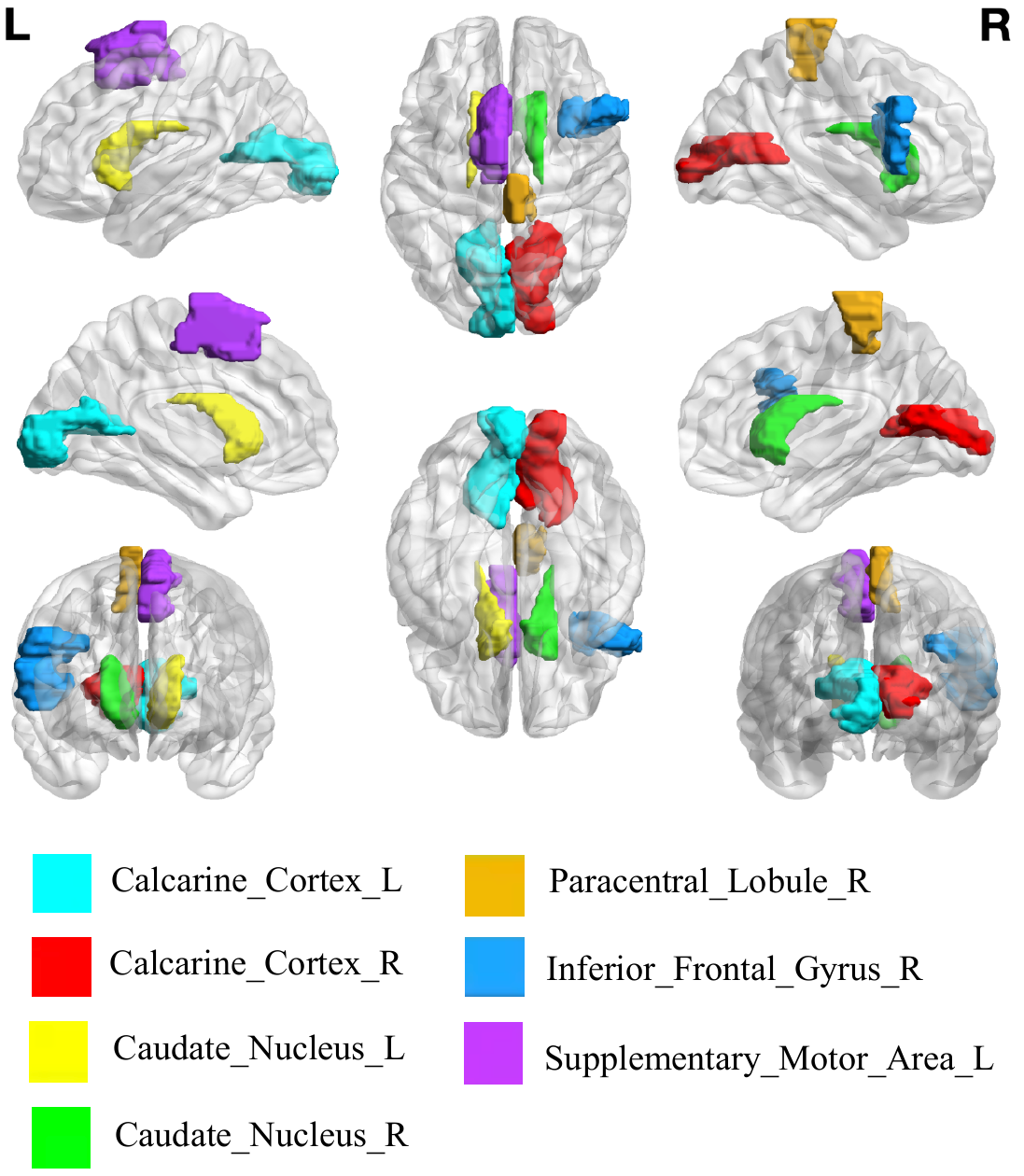}
    \caption{\textcolor{black}{multi-cohort SVM}}
    \label{fig:multi-cohort SVM}
    \end{subfigure}
    \caption{SVM training on the single-cohort dataset and multi-cohort dataset, result in 30\% and 35\% overlap cross dataset respectively. The overlap of BSNIP and UCLA is focused on Amygdala\_R, Calcarine\_L, Calcarine\_R, Postcentral\_L, Postcentral\_R, and Caudate\_R. 
    The overlap brain regions of multi-cohort training includes Frontal\_Inf\_Oper\_R, Supp\_Motor\_Area\_L, Paracentral\_Lobule\_R, Caudate\_L) and Caudate\_R,  Calcarine\_L, Calcarine\_R}
    \label{fig: single-cohort and multi-cohort SVM}
\end{figure*}

The 30\% overlap nodes of the top 20 important brain regions between the BSNIP dataset and the UCLA dataset in single-cohort training, including six brain regions: the right amygdala (Amygdala\_R), left calcarine sulcus (Calcarine\_L), right calcarine sulcus (Calcarine\_R), left postcentral gyrus (Postcentral\_L), right postcentral gyrus (Postcentral\_R), and right caudate nucleus (Caudate\_R) (see Fig. \ref{fig:single-cohort SVM}).  
Within the multi-cohort training, there are seven overlap regions,  35\% \textcolor{black}{overlap} between the training on BSNIP when tested on UCLA and the training on UCLA when tested on BSNIP (see Fig. \ref{fig:multi-cohort SVM}), including right Inferior frontal gyrus (Frontal\_Inf\_Oper\_R), left supplementary motor area (Supp\_Motor\_Area\_L), right Paracentral lobule (Paracentral\_Lobule\_R), left caudate nucleus (Caudate\_L) and Caudate\_R,  Calcarine\_L, Calcarine\_R.

The SVM model consistently highlights the brain regions Calcarine\_L, Calcarine\_R, and Caudate\_R in both single-cohort and multi-cohort training. 
The importance of the calcarine also shows high consistency with existing medical biomarker research. It is identical to the following result of the BrainIB++ model.

\subsection{BrainIB++}
After training on the BSNIP dataset and testing on the UCLA dataset, the Subgraph Generator can generate brain region subgraphs for all participants. To quantify the information content within each node, the subgraphs are averaged to derive a ranking indicating the probability of each node being selected by the model. This ranking delineates the likelihood of each node’s inclusion in the subgraph. We compared the top 20 nodes to ensure consistency with the SVM model. This alignment allows for a fair evaluation of node importance across models and ensures that the selected features are comparable in both approaches.

\begin{figure*}[h]
    \centering
    \begin{subfigure}[b]{0.31\linewidth}
    \centering
    \includegraphics[width=1\linewidth]{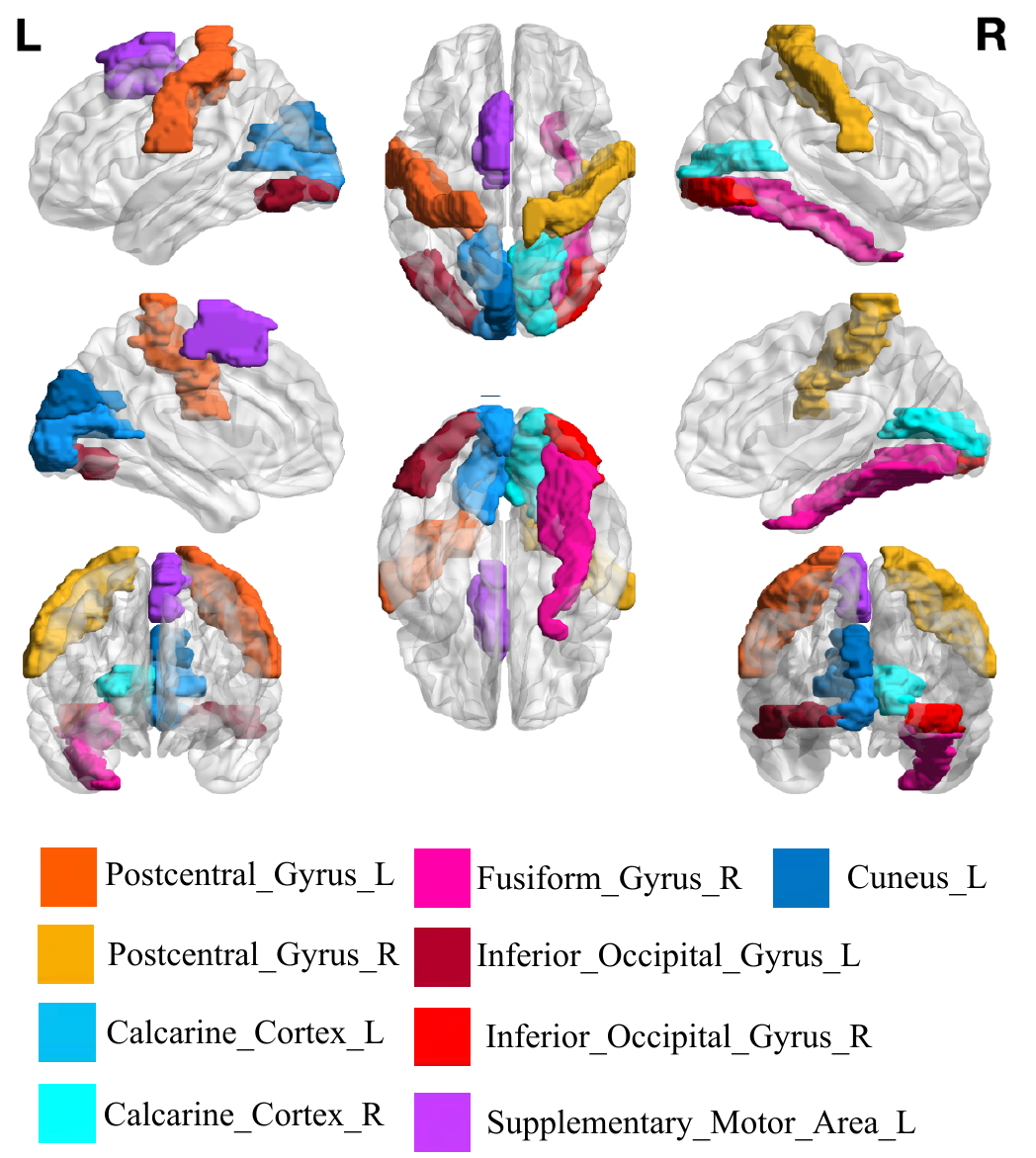}
    \caption{\textcolor{black}{BSNIP dataset}}
    \label{fig:BSNIP_nodes}
    \end{subfigure}
         \hfill
    \begin{subfigure}[b]{0.355\linewidth}
    \centering
    \includegraphics[width=1\textwidth]{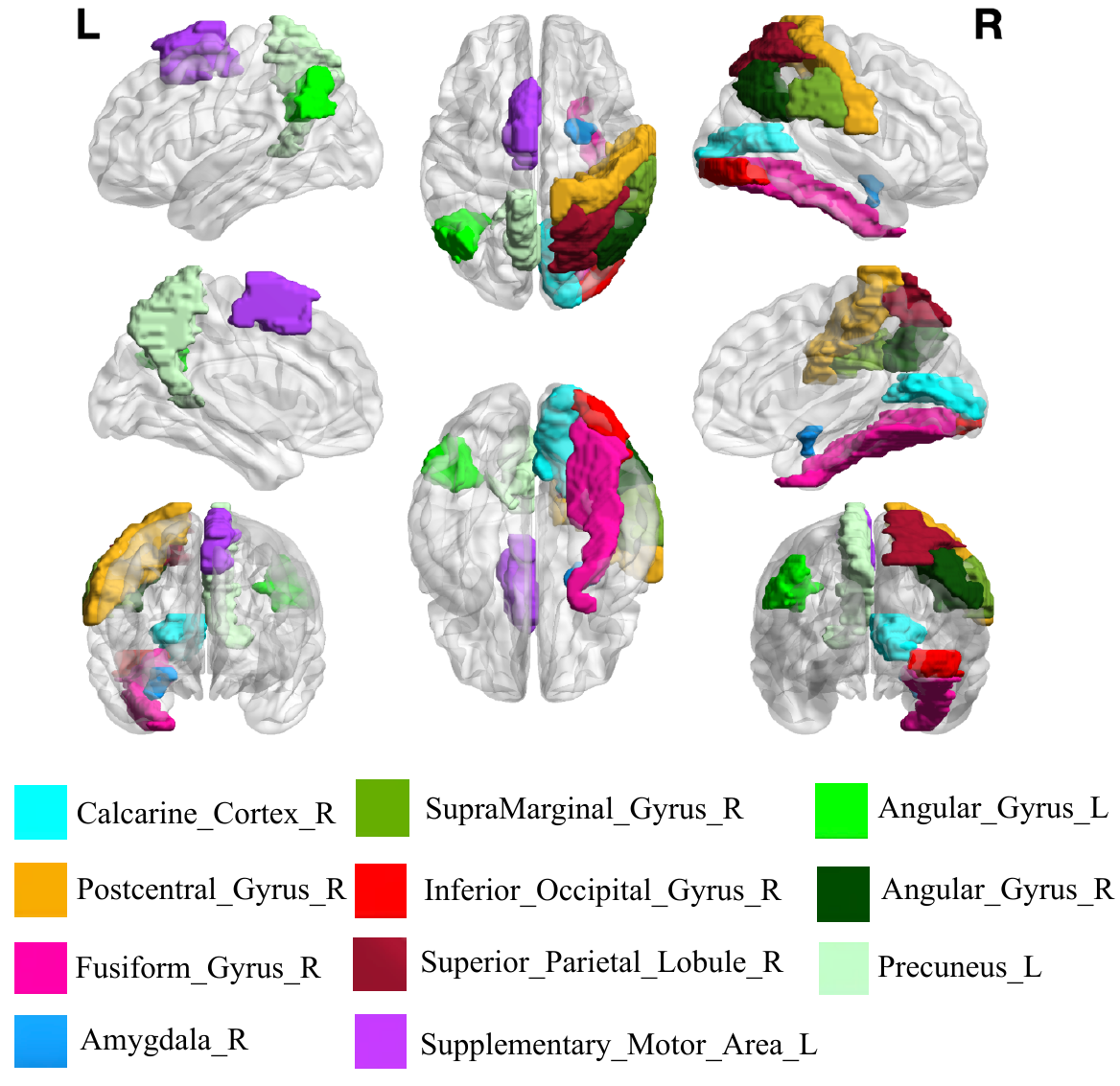}
    \caption{\textcolor{black}{UCLA dataset}}
    \label{fig:UCLA_nodes}
    \end{subfigure}
    \hfill
    \begin{subfigure}[b]{0.31\linewidth}
    \centering
    \includegraphics[width=1\textwidth]{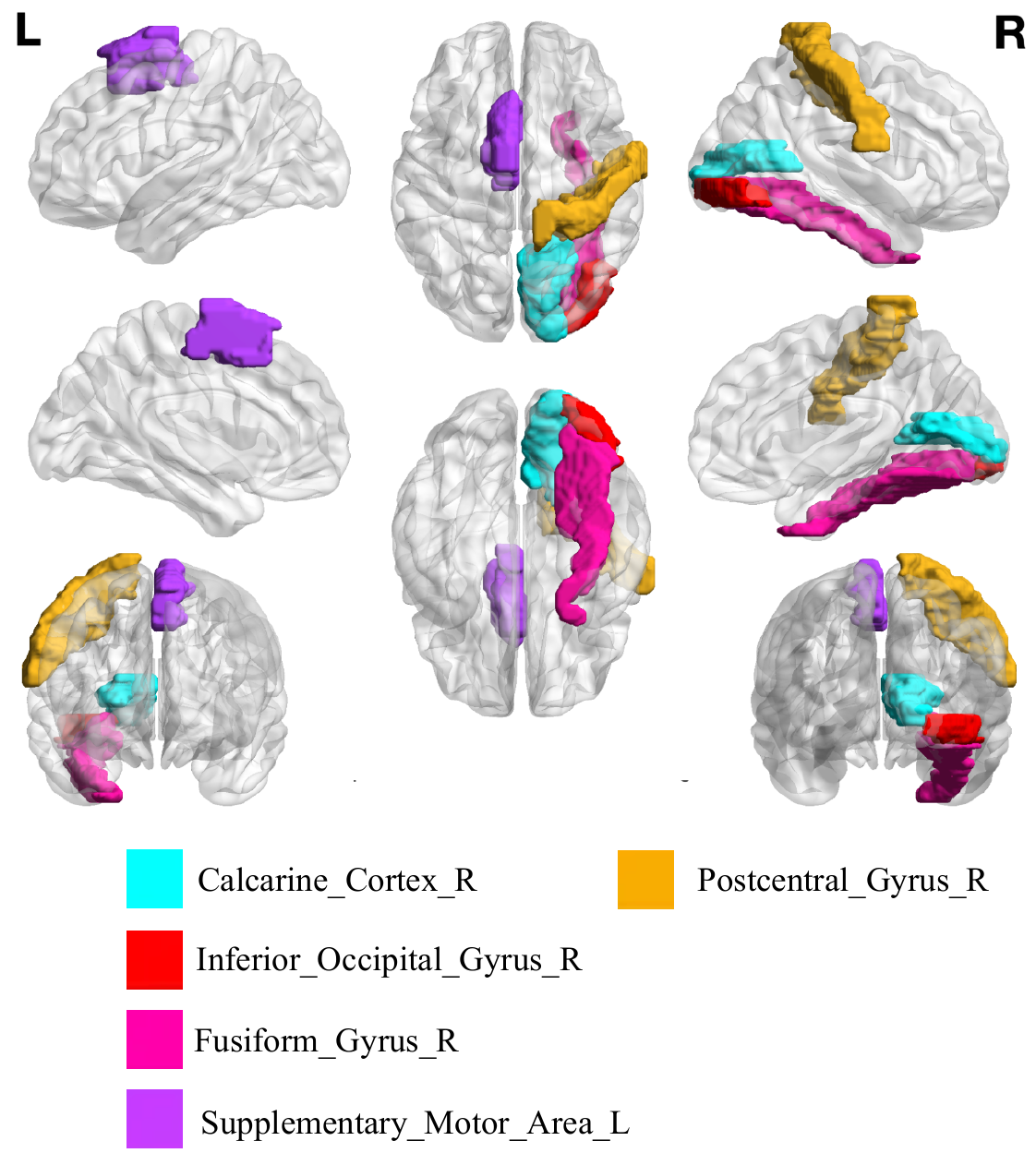}
    \caption{\textcolor{black}{overlap on BSNIP and UCLA dataset}}
    \label{fig: multi-cohort BrainIB++ Overlap}
    \end{subfigure}
    \caption{The BrainIB++ model demonstrated 40\% overlap in distinct node preferences across both the BSNIP and UCLA datasets during multi-cohort training. Fig.(a) and Fig.(b) illustrate the nodes with the highest probabilities selected by the well-trained BrainIB++ subgraph generator for the BSNIP dataset and the UCLA dataset, respectively. 
    Fig.(c) displays the five common nodes identified in both (a) and (b), including Supp\_Motor\_Area\_L, Calcarine\_R, Occipital\_Inf\_R, Fusiform\_R, Postcentral\_R. Supp\_Motor\_Area\_L and Calcarine\_R appeared more than once  within the overlapping nodes.}
    \label{fig: multi-cohort BrainIB++}
\end{figure*}

By inputting the BSNIP and UCLA datasets into the subgraph generator trained on cross-cohort data (BSNIP to UCLA), the BrainIB++ model demonstrated a 40\% overlap in node preferences between the two datasets. This overlap highlights the model’s ability to identify consistent and robust brain region biomarkers across cohorts, despite differences in data distributions.
It demonstrated a strong preference for certain node clusters, including right inferior occipital lobe (Occipital\_Inf\_R), right fusiform (Fusiform\_R), right postcentral gyrus (Postcentral\_R), Supp\_Motor\_Area\_L and Calcarine\_R (see Fig. \ref{fig: multi-cohort BrainIB++}). Notably, Supp\_Motor\_Area\_L and Calcarine\_R appeared multiple times within the overlapping nodes, emphasizing their significance across the datasets. This finding indicates that specific nodes are integral to the model’s classification decisions.  
Our model consistently includes these nodes in the subgraphs to minimize loss, suggesting that they contain compressed, relevant information associated with schizophrenia. Furthermore, the overlap observed across both datasets underscores the model’s robust generalization capabilities across data from multiple cohorts. 

BrainIB++ and SVM are the only two models that demonstrate generalization capability. SVM’s ability to generalization from its reliance on manually crafted features, which are inherently designed to be robust across datasets. However, the creation of such features is often labor-intensive and time-consuming. In contrast, BrainIB++ achieves generalizability by removing noisy and irrelevant nodes, enabling it to adapt effectively to new data.

\subsection{Statistical Analysis}
\textcolor{black}{To investigate potential differences in brain network organization between patient and control groups, we conducted statistical analyses on graph-based connectivity features. Given the limited sample sizes of the UCLA and COBRE datasets, topological analysis was performed exclusively on the larger BSNIP dataset to ensure statistical robustness.}
\textcolor{black}{For completeness and validation purposes, equivalent analyses of the UCLA and COBRE datasets are documented in \textit{supplementary file}.}
\textcolor{black}{For each subject, the brain network was constructed following the same preprocessing pipeline described in the experimental section, resulting in a graph where nodes represent brain regions and edges denote functional connectivity. From each graph, we extracted global-level topological metrics including average degree, 
average clustering coefficient, 
and global efficiency. 
At the nodal level, we computed node-level topological metrics such as nodal degree, nodal efficiency, and betweenness centrality. Specifically, given a graph $\mathbf{G}=\{\mathcal{V};\mathcal{E}\}$. $\mathcal V = \{ V_i|i \in \{1,...,n \} \}$ denote nodes with node feature matrix $X$ and $n$ is the number of nodes.}

\textcolor{black}{
For a node $V_i$ in graph $G$, its degree is: 
\begin{align}
    \text{Degree}_{nodal}(V_i) = \sum_{j \in \mathcal V} A_{ij}, 
\end{align} 
Where $A$ is the adjacency matrix, $A_{ij} = 1$ means that node $V_i$ is connected to node $V_j$. 
The efficiency of a node $V_i$ is defined as the average of the sum of the reciprocals of the shortest path lengths between it and all other nodes:
\begin{align}
    \text{Efficiency}_{nodal}(V_i) = \frac{1}{|n|-1} \sum_{j \in \mathcal V, j \ne i} \frac{1}{d_{ij}}, 
\end{align}
where $d_{ij}$ denotes the shortest path length from $V_i$ to $V_j$. 
The betweenness centrality of node $V_i$ is:
\begin{align}
        \text{Centrality}_{nodal}(V_i) = \sum_{V_s \ne V_i \ne V_t \in \mathcal V} \frac{\sigma_{st}(V_i)}{\sigma_{st}},
\end{align}
where $\sigma_{st}$ is the number of shortest paths from node $V_s$ to node $V_t$, and $\sigma_{st}(V_i)$ is the number of these paths that pass through the node $V_i$.
Furthermore, global degree and efficiency were derived by aggregating the corresponding nodal-level metrics:
\begin{align}
    \text{Degree}_{global} = \frac{1}{n} \sum_{V_i \in \mathcal V} \sum_{j \in \mathcal V} A_{ij}, \\
    \text{Efficiency}_{global} = \frac{1}{n (|n|-1)} \sum_{V_i \in \mathcal V} 
     \sum_{j \in \mathcal V, j \ne i} \frac{1}{d_{ij}},
\end{align}
We employ the clustering coefficient rather than the average betweenness centrality as a metric. This choice is motivated by the fact that average betweenness centrality is frequently influenced disproportionately by a small subset of highly centralized nodes, resulting in limited stability and representativeness when characterizing the overall network. Average betweenness is more appropriate for identifying critical hub nodes rather than assessing the integrative properties or local integration of the entire network. Conversely, the average clustering coefficient quantifies the extent to which information is propagated and redundantly distributed within local neighborhoods. As such, it serves as a robust indicator of the network’s local integration capacity and is widely adopted to characterize the local connectivity patterns (i.e., local segregation) of complex networks at a global scale.
A high value means that the network is more locally dense, and neighbors of nodes tend to be interconnected.
The global network average clustering coefficient is the average of all nodes' average clustering coefficient.
\begin{align}
    Clustering_{global} = \frac{1}{n} \sum_{V_i \in \mathcal V} \frac{2 \times \text{N}_{V_i}}{\text{Degree}_{V_i} \times (\text{Degree}_{V_i} - 1)} ,
\end{align}
where $\text{N}_{V_i}$ is the number of links between neighbors of node $V_i$.
}

\textcolor{black}{We conducted a comprehensive comparison of both node-level and global topological metrics between schizophrenia (SZ) patients and healthy controls (HC) to assess network-level alterations. }

\textcolor{black}{At the global level (Table \ref{Statistic_Analysis_global}), only global efficiency exhibited a statistically significant difference between groups (T = -2.451, p = 0.014), with SZ patients showing reduced efficiency. This finding suggests impaired capacity for parallel information transfer across the whole brain network, consistent with the hypothesis of disrupted functional integration in schizophrenia. In contrast, the average degree (p = 0.377) and average clustering coefficient (p = 0.426) showed no significant group differences, indicating that basic connection density and local clustering patterns remain largely preserved.}

\begin{table}[width=.6\linewidth,cols=4,pos=h]
\centering
\caption{\textcolor{black}{Group comparison of global topological metrics between patients and controls in BSNIP. Bold p-values indicate statistical significance.}}
\label{Statistic_Analysis_global}
{\color{black} 
\begin{tabular}{lccc}
\toprule
\textbf{Metric} & \textbf{T-value} & \textbf{p-value} & \textbf{Significance} \\
\midrule
Average Degree         & -0.884 & 0.377 &  Not significant \\
Average Clustering     &  0.795 & 0.426 &  Not significant \\
Global Efficiency      & \textbf{-2.451} &  \textbf{0.014} &\textbf{Significant} \\
\bottomrule
\end{tabular}
}
\end{table}

\begin{table}[width=1\linewidth,pos=h]
\centering
\caption{\textcolor{black}{Group comparison of topological metrics between patients and controls in BSNIP. Bold p-values indicate statistical significance.}}
\label{Statistic_Analysis_node}
\color{black}{
\begin{tabular}{cccccccccc}
\toprule
\multirow{2}{*}{\textbf{BrianIB++}} 
& \multicolumn{2}{c}{\textbf{Degree}} 
& \multicolumn{2}{c}{\textbf{Efficiency}} 
& \multicolumn{2}{c}{\textbf{Betweenness}} 
& \multicolumn{3}{c}{\textbf{p-value}} \\
\cline{2-10} \\
[-0.3cm]
 & HC & SZ & HC & SZ & HC & SZ & Degree & Efficiency & Betweenness \\
\midrule
Cuneus\_L & 5.764 & 5.89 & 0.601 & 0.610 & 0.028 & 0.031 & 0.461 & 0.571 & 0.124 \\
Fusiform\_R & 12.458 & 12.976 & 0.771 & 0.756 & 0.032 & 0.037 & \textbf{0.034} & \textbf{0.012} & \textbf{0.028} \\
Occipital\_Inf\_L & 9.969 & 9.886 & 0.821 & 0.804 & 0.016 & 0.017 & 0.692 & \textbf{0.016} & 0.371 \\
Occipital\_Inf\_R & 9.695 & 9.548 & 0.820 & 0.817 & 0.015 & 0.015 & 0.464 & 0.709 & 0.937 \\
Supp\_Motor\_Area\_L & 9.567 & 9.375 & 0.740 & 0.758 & 0.019 & 0.019 & 0.444 & 0.102 & 0.741 \\
Calcarine\_R & 9.888 & 9.667 & 0.726 & 0.725 & 0.034 & 0.037 & 0.376 & 0.944 & 0.299 \\
Postcentral\_L & 8.873 & 8.110 & 0.772 & 0.732 & 0.014 & 0.015 & \textbf{0.007} & \textbf{0.005} & 0.427 \\
Postcentral\_R & 8.961 & 8.217 & 0.762 & 0.739 & 0.015 & 0.015 & \textbf{0.010} & 0.117 & 0.952 \\
[0.1cm]
\toprule
\multirow{2}{*}{\textbf{SVM}} 
& \multicolumn{2}{c}{\textbf{Degree}} 
& \multicolumn{2}{c}{\textbf{Efficiency}} 
& \multicolumn{2}{c}{\textbf{Betweenness}} 
& \multicolumn{3}{c}{\textbf{p-value}} \\
\cline{2-10} \\
[-0.3cm]
 & HC & SZ & HC & SZ & HC & SZ & Degree & Efficiency & Betweenness \\
\midrule
Amygdala\_R        & 11.322 & 11.428 & 0.794 & 0.802 & 0.029 & 0.027 & 0.579 & 0.220 & 0.317 \\
Caudate\_R         & 9.612  & 9.861  & 0.858 & 0.862 & 0.010 & 0.010 & 0.339 & 0.705 & 0.990 \\
Calcarine\_L       & 8.758  & 8.916  & 0.790 & 0.791 & 0.015 & 0.016 & 0.496 & 0.911 & 0.600 \\
Calcarine\_R       & 9.888  & 9.667  & 0.726 & 0.725 & 0.034 & 0.037 & 0.376 & 0.944 & 0.299 \\
Postcentral\_L     & 8.873  & 8.110  & 0.772 & 0.732 & 0.014 & 0.015 & \textbf{0.007} & \textbf{0.005} & 0.427 \\
Postcentral\_R     & 8.961  & 8.217  & 0.762 & 0.739 & 0.015 & 0.015 & \textbf{0.010} & 0.117 & 0.952 \\
\bottomrule
\end{tabular}
}
\end{table}

\textcolor{black}{Table \ref{Statistic_Analysis_node}
present comparisons of node topological metrics (Degree, Efficiency, and Betweenness) between HC and SZ across several brain regions selected by the BrainIB++ and SVM model. Each metric reflects a distinct aspect of network function: degree indicates the number of direct connections, efficiency captures local information transfer, and betweenness reflects a node’s role in global communication.}

\textcolor{black}{
Specifically, the degree quantifies how connected a node is within the network. In the BrainIB++ model, two regions exhibited significant group differences. Postcentral\_L (p = 0.007) and Postcentral\_R (p = 0.010) both showed reduced degrees in SZ, suggesting decreased direct connectivity in bilateral somatosensory cortices. This aligns with altered bodily sensation and tactile processing frequently reported in schizophrenia. The SVM model also identified the same two regions (Postcentral\_L and Postcentral\_R) as significant.
}

\textcolor{black}{Local efficiency reflects the ease of communication within a node’s neighborhood, indicating fault tolerance and integration capacity. BrainIB++ revealed three regions with significant group differences: Fusiform\_R (p = 0.012), Occipital\_Inf\_L (p = 0.016) and Postcentral\_L (p = 0.005).
The lower efficiency of Fusiform\_R in SZ may reflect impaired visual integration, especially in face and object recognition. In the meantime, reduced efficiency of Occipital\_Inf\_L may point to dysfunctional early visual processing. The result in Postcentral\_L confirms the findings from the degree analysis, this result further supports locally impaired information transfer in the somatosensory network.
In contract, SVM detected only Postcentral\_L as significant (p = 0.005), which indicating its limited detection capacity.}

\textcolor{black}{Betweenness centrality captures how often a node lies on the shortest path between other nodes, reflecting its importance in global communication. BrainIB++ identified the Fusiform\_R (p = 0.028) as a significant region. This region showed higher betweenness in SZ, suggesting it may act as a compensatory communication hub, despite its reduced local efficiency. Such a shift might indicate a reorganization of the visual network hierarchy under pathological conditions.
However, no regions showed significant betweenness differences in the SVM results, highlighting the advantage of BrainIB++ in uncovering global-level network alterations.
}

\textcolor{black}{
By analyzing three complementary topological metrics, BrainIB++ uncovered four distinct brain regions with significant alterations in SZ, across different levels of network organization. These included the fusiform gyrus and inferior occipital cortex (visual processing), and bilateral postcentral gyri (somatosensory function), reflecting both local and global dysconnectivity patterns. In contrast, the SVM model identified only the postcentral regions, suggesting reduced sensitivity and a narrower detection scope.
These results highlight BrainIB++’s capacity to reveal a richer and more comprehensive profile of schizophrenia-related network disruptions, through interpretable and biologically meaningful node-level metrics.
}

\textcolor{black}{For completeness, we report the results of statistical tests conducted across all nodal and global metrics, with Benjamini-Hochberg procedure as false discovery rate (FDR) corrections, in the Supplement File.
}

\section{Discussion}

\subsection{Comparison of ML models and GNN}
In comparing the performance of our model with baseline approaches, SVM was expected to exhibit the best performance among the machine learning methods. However, this approach yielded disappointing classification results with the generated hand-crafted features on the single-cohort training. 
\textcolor{black}{All models exhibited trends of overfitting on the UCLA and COBRE datasets, likely due to its significantly smaller size compared to other datasets. Conversely, the Decision Tree and AdaBoost models demonstrated underfitting. Additionally, Graph Transformer required over 15 times more training time than the other GNNs, making it impractical for inclusion in the final BrainIB++ model.}

Moreover, during the interpretable analysis, while the number of overlapping nodes between SVM and BrainIB++ was similar, BrainIB++ achieved a higher overlap percentage (40\% compared to SVM’s 30\% and 35\%). This demonstrates that BrainIB++ not only identifies comparable key nodes but also exhibits greater consistency and robustness in capturing shared brain region features across datasets.

Consequently, BrainIB++ demonstrates superior performance in both classification accuracy and built-in interpretability. By combining efficient computation with the ability to generate clear, detailed explanations, BrainIB++ addresses the limitations of traditional machine learning approaches, offering a robust and transparent solution for complex neuroimaging tasks.

\subsection{Biomarkers for Schizophrenia}

In our investigation of biomarkers, several brain region clusters consistently emerged, aligning with the current literature. Notably, the right calcarine, right fusiform, inferior occipital lobe and supplementary motor area are recognized as important biomarkers for schizophrenia in both our research and prior studies.  Our study also identified the right postcentral \textcolor{black}{gyrus} as a critical factor in schizophrenia detection, an area that has not been as thoroughly explored as others.

Specifically, the Calcarine\_R, Fusiform\_R, and the Occipital\_Inf\_R play significant roles in visual information processing, including object recognition, face recognition, and facial expression recognition \cite{weiner2016anatomical}.
\textcolor{black}{These brain regions have been consistently associated with clinical symptoms such as visual hallucinations, impaired facial emotion perception, and social cognition deficits.}
Koshiyama et al. reported that resting-state network abnormalities were detected in the frontal lobe and near the fusiform gyrus and calcarine sulcus in patients with schizophrenia \cite{koshiyama2020neurophysiologic}.
\textcolor{black}{These abnormalities suggest over-synchronized activation in perception-related brain regions, potentially contributing to hallucinations, abnormal salience attribution, and delusions.}

The inferior occipital lobe serves as the visual processing center of the mammalian brain and contains the majority of the anatomical area of the visual cortex. Schizophrenia is associated with alterations in the occipital lobe, including changes in both gray and white matter volume. Variations in the volume of the occipital lobe have also been associated with schizophrenia \cite{tohid2015alterations}. 
\textcolor{black}{Occipital bending and structural abnormalities in this region have been linked to visual perceptual disturbances and hallucinations \cite{maller2017occipital, anderson2013overlapping}. 
The occipital bending is more prevalent among patients with schizophrenia than among healthy subjects, patients with schizophrenia showed significantly lower fractional anisotropy in temporal and occipital white matter compared to healthy volunteers.
Moreover, schizophrenia patients exhibit reduced persistence of visual sensory information, and functional connectivity disruptions in the occipital region have been observed even in early-stage patients \cite{uscuatescu2021reduced, white2011disrupted}.}

The fusiform gyrus (FG) is regarded as a key structure for specialized computations of higher-level visual functions, including face perception, object recognition, and reading \cite{weiner2016anatomical}. Schizophrenia is associated with a bilateral reduction in gray matter volume within the fusiform gyrus \cite{lee2002fusiform}. Patients with chronic schizophrenia exhibited overall smaller FG gray matter volumes (10\%) than normal controls. Additionally, these patients performed more poorly than normal controls on both immediate and delayed facial memory tests. The degree of impairment on delayed memory for faces was significantly correlated with the extent of bilateral anterior FG reduction in patients with schizophrenia \cite{onitsuka2003fusiform}.

\textcolor{black}{Furthermore, our model identified the right postcentral gyrus and supplementary motor area as relevant to schizophrenia. The postcentral gyrus, part of the primary somatosensory cortex, processes tactile and proprioceptive input.}
Liang et al. reported a decrease in hemodynamic response in the right postcentral cortex, which is associated with treatment-resistant auditory verbal hallucinations in schizophrenia \cite{liang2022decrease}. 
\textcolor{black}{This suggests that disrupted sensory feedback mechanisms may lead to the misattribution of internal speech as external voices.}
Cai et al. similarly found a significant negative correlation between altered local functional connectivity and illness duration in the right precentral and postcentral gyrus \cite{cai2022disrupted}.
\textcolor{black}{The supplementary motor area plays a key role in voluntary movement and motor planning.}
\textcolor{black}{Altered activity in this region has been linked to psychotic motor symptoms such as catatonia, stereotyped movement, and reduced initiation of motor behavior \cite{stegmayer2014supplementary, jankovic2021principles}. }
Schröder et al. employed functional fMRI to investigate the activation of the sensorimotor cortex and supplementary motor area during finger-to-thumb opposition, finding that patients with schizophrenia exhibited decreased activation in both sensorimotor cortices and the supplementary motor area, as well as a reversed lateralization effect compared to healthy controls \cite{schroder1995sensorimotor}.

\textcolor{black}{Overall, BrainIB++ has effectively identified both previously recognized and novel biomarkers. By linking these neurobiological features to specific clinical symptoms, our findings provide a more comprehensive understanding of the neural mechanisms underlying schizophrenia and support the potential of interpretable models in clinical neuroscience research.}

\section{Conclusion and Future work}
In this study, we aimed to utilize artificial intelligence methods to identify critical brain regions associated with schizophrenia. To achieve this, we developed a classification model named BrainIB++, which is grounded in graph neural networks and the information bottleneck theory. By comparing mutual information values between the sampled subgraph and the original graph data, and incorporating these graphs into the classification process, the well-trained Subgraph Generator module identifies the most informative subgraph for a new subject. This subgraph is deemed to contain the most pertinent information for schizophrenia, representing the most critical brain regions.

The BrainIB++ model exhibits superior accuracy compared to traditional machine learning models and SOTA models. To further evaluate its performance, we tested the model on multi-cohort datasets to assess its built-in interpretability and generalization capabilities, conducted a quantitative analysis of its accuracy. BrainIB++ also outperforms traditional models like SVM with a higher percentage of overlap.
The results indicate that our model focuses on specific brain regions, including the left supplementary motor area, right calcarine, right inferior occipital, right fusiform, and right postcentral. These findings align with results obtained from traditional hand-crafted models and are supported by clinical evidence.

Although BrainIB++ effectively balances complex preprocessing and feature engineering with post-hoc interpretability, the current model is not yet suitable for clinical application due to the need for larger datasets to achieve more accurate classification results. To overcome the limitations posed by limited data, future research could explore the integration of multi-modal data, which holds promise for enhancing the model’s performance and generalizability.

\textcolor{black}{Furthermore, this study focused solely on the importance of nodes (i.e., brain regions), while the contributions of edges (i.e., functional connections between brain regions) were not explicitly considered. However, functional connectivity is known to play a crucial role in shaping neural dynamics and is often altered in various brain disorders. Ignoring edge-level information may therefore limit the interpretability and biological relevance of the identified subgraphs. As a direction for future work, we plan to develop a more comprehensive subgraph generation mechanism that jointly learns node and edge importance. By incorporating both topological and functional aspects of the brain network, this integrated approach is expected to enhance the quality and faithfulness of the resulting explanations.}

\printcredits

\section*{Declaration of Competing Interest}
\small The authors declare that they have no known competing financial interests or personal relationships that could have appeared to influence the work reported in this paper.

\section*{Funding}
\small This research did not receive any specific grant from funding agencies in the public, commercial, or not-for-profit sectors.

\section*{Data Availability}
\small \textcolor{black}{The UCLA, COBRE and BSNIP datasets are publicly available (\url{https://exhibits.stanford.edu/data/catalog/mg599hw5271}, \url{http://fcon_1000.projects.nitrc.org/indi/retro/cobre.html}, \url{http://b-snip.org/}).} 
The preprocessed data can be found in \url{ https://drive.google.com/drive/folders/1ca9-nxsldpN3Cam_bGX6odce1VlOt96J?usp=share_link}.

\section*{Code Availability}
\small \textcolor{black}{The code of our BrainIB++ is available at \url{https://github.com/TianzhengHU/BrainIB_coding/tree/main/BrainIB_GIB}}.




\bibliographystyle{unsrt}
\bibliography{references}

\end{document}